\documentclass[final,5p,times,twocolumn]{elsarticle}

\usepackage{graphics}
\usepackage{amsmath}
\usepackage{color}
\usepackage{marvosym}
\usepackage{multirow} 
\usepackage{booktabs}
\usepackage{bm}
\usepackage[switch]{lineno}
\usepackage{makecell}
\usepackage{url}
\journal{Pattern Recognition}
\newcolumntype{L}[1]{>{\raggedright\arraybackslash}p{#1}}
\newcolumntype{C}[1]{>{\centering\arraybackslash}p{#1}}
\newcolumntype{R}[1]{>{\raggedleft\arraybackslash}p{#1}}
\newcommand\clr[1]{{\color{black}{#1}}}

\usepackage{amsmath,amsfonts,bm}

\def\eqref#1{equation~\ref{#1}}

\def\1{\bm{1}}

\DeclareMathAlphabet{\mathsfit}{\encodingdefault}{\sfdefault}{m}{sl}
\SetMathAlphabet{\mathsfit}{bold}{\encodingdefault}{\sfdefault}{bx}{n}

\newcommand{\R}{\mathbb{R}}

\usepackage{amsthm}

\newtheorem{definition}{Definition}

\theoremstyle{plain}

\newtheorem{proposition}{Proposition}

\usepackage[utf8]{inputenc} \usepackage[T1]{fontenc}    \usepackage{hyperref}       \usepackage{amsfonts}       \usepackage{nicefrac}       \usepackage{microtype}      \usepackage{xcolor}         

\usepackage{graphicx}
\usepackage[normalem]{ulem}
\useunder{\uline}{\ul}{}
\usepackage{adjustbox}

\usepackage {diagbox}
\usepackage{wasysym}
\usepackage{caption}

\usepackage{balance}

\begin{document}
\begin{frontmatter}

\title{MambaTS: Improved Selective State Space Models for Long-term Time Series Forecasting}

\author[1,2]{Xiuding Cai}
\author[1,2]{Xueyao Wang}
\author[1,2]{Yaoyao Zhu}
\author[1,2]{Yu Yao}

\affiliation[1]{organization={Chengdu Institute of Computer Application, Chinese Academy of Sciences},city={Chengdu},
	postcode={610213}, 
country={China}}
\affiliation[2]{organization={University of Chinese Academic Sciences},city={Beijing},
	postcode={101408}, 
country={China}}

\begin{abstract}
\clr{
	In recent years, Transformers have become the de-facto architecture for long-term time series forecasting (LTSF), yet they face challenges associated with the self-attention mechanism, including quadratic complexity and permutation-invariant bias. This raises an important question: \emph{do we truly need self-attention to model long-range dependencies in LTSF?} To address this, we propose MambaTS, a linear-scan-based framework that models global dependencies across time and variables via structured dependency modeling. Since explicit variable dependency structures are often unknown, we introduce Variable-Aware Scan along Time (VAST), which learns inter-variable relationships during training and determines an optimal scan order via a shortest-path-based decoding strategy during inference. MambaTS employs the latest Mamba model as its backbone. We suggest that the causal convolution in the vanilla Mamba is unnecessary due to the presence of independent variables, leading to the development of the Temporal Mamba Block (TMB). To mitigate model overfitting, we further incorporate a dropout mechanism for selective parameters in TMB. Extensive experiments conducted on eight public datasets demonstrate that MambaTS achieves competitive or state-of-the-art performance on most datasets. Code is available at this repository: \href{https://github.com/XiudingCai/MambaTS-pytorch}{https://github.com/XiudingCai/MambaTS-pytorch}.
}
\end{abstract}

\end{frontmatter}

\section{Introduction}
Long-term time series forecasting (LTSF) has a wide range of applications in various fields, including weather~\cite{wu2023interpretable}, finance~\cite{shi2026stock}, healthcare~\cite{ye2026medspaformer}, energy~\cite{wu2021autoformer} and transportation~\cite{lv2025multimodal}. With the rapid advancement of deep learning, the current methods for time series prediction have shifted from traditional statistical learning approaches to deep learning-based methods~\cite{qiu2024tfb}, such as recurrent neural networks (RNNs;~\cite{lin2026segrnn}) and temporal convolutional neural networks (TCNs;~\cite{zhang2025dynamic}). Since the introduction of Transformer~\cite{vaswani2017attention}, Transformer-based methods have emerged as the mainstream LTSF approach~\cite{zhou2021informer, nie2023a, ren2023temporal}, leveraging their self-attention mechanism to effectively capture long-term dependencies in time series data. 

Recent studies have identified critical challenges associated with the application of Transformers in LTSF, primarily attributed to the self-attention mechanism inherent in Transformers. One major concern is that the self-attention mechanism often suffers from the curse of quadratic complexity, resulting in a computational cost that escalates rapidly with the context length~\cite{wu2021autoformer, zhou2022fedformer}. Furthermore, recent research has shown that the performance of Transformer-based LTSF methods does not necessarily improve with an increasing look-back window~\cite{nie2023a, liu2024itransformer}. This phenomenon may be due to the distracted attention caused by the growing input size~\cite{liu2024itransformer}. Additionally, a recent study titled DLinear~\cite{Zeng2023Dlinear} has challenged the effectiveness of the permutation-invariant bias of the self-attention mechanism in LTSF, achieving remarkable results that surpass those of most state-of-the-art (SOTA) Transformer-based methods using a simple single-layer feed-forward network.

This prompts us to question: \emph{do we truly need the self-attention mechanism to establish long-range dependencies in LTSF?} To address this, we can discuss the issue in two parts. First, regarding temporal dependencies, since time series data is inherently causal, a single linear complexity is sufficient for modeling. Second, when considering variable dependencies, do we require pairwise computations to effectively model these relationships? The answer is no.

Let us establish intuition through a simple example. Consider a causal graph with directed edges representing dependencies among variables, where we aim to predict future outcomes based on these relationships. The causal graph can be represented with edges \(A \rightarrow C\) and \(B \rightarrow C\). Due to the nature of causality, a linear scan such as \(A, B, C\) or \(B, A, C\) effectively captures global dependencies. This is because the information from both \(A\) and \(B\) inherently contributes to predicting \(C\), thus eliminating the need for complex pairwise calculations.

Based on this insight, we propose MambaTS, a novel architecture for LTSF that models global dependencies with linear complexity. MambaTS is motivated by the observation that an appropriate variable ordering can facilitate efficient dependency modeling through a single linear scan.

However, the underlying dependency structure among variables is generally unknown in practice. To address this challenge, we introduce Variable-Aware Scan along Time (VAST). During training, VAST estimates inter-variable dependencies through a random walk without return strategy, while during inference it employs a shortest-path-based decoding method~\cite{dreo2006metaheuristics} to determine an effective linear scan order.

%

We utilize the latest Mamba~\cite{gu2023mamba} model as our encoder network. As a formidable competitor to Transformer architectures, Mamba enhances traditional state space model (SSM)~\cite{fu2022h3, gu2021efficiently, zhang2023spacetime} by introducing selection mechanisms to filter out irrelevant information and reset states along. It also incorporates hardware-aware design for efficient parallel training. Mamba has demonstrated competitive performance compared to Transformers across various domains~\cite{liu2024vmamba, li2024videomamba, liang2024pointmamba}, offering rapid inference and scalability with sequence length.

However, we note that the causal convolution in the original Mamba block can be detrimental when scanning independent variables. Consequently, we have removed the local convolution prior to the SSM, leading to the introduction of the Temporal Mamba Block (TMB). Furthermore, previous studies have shown that excessive information integration can lead to overfitting~\cite{nie2023a}, a phenomenon we also observed during experiments with MambaTS. To mitigate this, we have added a dropout mechanism~\cite{srivastava2014dropout} for selective parameters in TMB.

Gained from the insights and designs, MambaTS achieves efficient modeling of global dependencies across time and variables with linear complexity. Extensive experiments on eight popular public datasets demonstrate that MambaTS achieves SOTA performance cross most LSTF datasets and settings.

Our contributions are as follows:
\begin{enumerate}
	\item We introduce MambaTS, a new time series forecasting model based on selective state space models, achieving global dependency modeling through linear scans, supported by theoretical evidence.
	\item We propose a method for estimating inter-variable dependencies during training via random walks without return, and a decoding strategy based on shortest-path optimization to determine an effective linear scan sequence during inference.
	\item We present the Temporal Mamba Block, which avoids entanglement between independent channels by removing the original causal convolution. Additionally, we incorporate a dropout mechanism for selective parameters in TMB to further prevent overfitting.
	\item Our experimental results demonstrate competitive or SOTA across most datasets and settings, especially on high-dimensional and complex multivariate datasets.
\end{enumerate}

\section{Related Work}
\label{related_work}

\paragraph{Long-Term Time Series Forecasting}
Traditional LTSF methods leverage the statistical properties and patterns of time series data for prediction. In recent years, LTSF has shifted towards deep learning approaches, where various neural networks are utilized to capture complex patterns and dependencies, elevating LTSF performance~\cite{zhou2021informer, wu2021autoformer, kim2025comprehensive}. These methods can be broadly classified into two categories: variable-mixing and variable-independent. Variable-mixing methods employ diverse architectures to model dependencies across time and variables. RNNs~\cite{lai2018modeling, lin2026segrnn} were initially introduced to LTSF due to the nature of sequence modeling. TCNs, known for their local bias, are effective in capturing local patterns in time series data and have shown promising results in LTSF~\cite{wu2023timesnet, gu2025hdtcnet}. Transformers are subsequently introduced to accomplish long-range dependency modeling through self-attention and have become a mainstream method in the Transformer family~\cite{wen2022transformers}. However, due to the quadratic complexity, Transformer-based methods have struggled with optimization efficiency~\cite{zhou2022fedformer, wei2025wavelet}. Recently, significant improvements have been made in these methods with the introduction of patch-based techniques~\cite{nie2023a, ling2025eapformer}. MLPs are also commonly used for LTSF and have achieved impressive results with their simple and direct architectures~\cite{wang2025timemixer++}. Graph neural networks have been utilized to model relationships between variables~\cite{wu2020mtgnn}. FourierGNN~\cite{yi2023fouriergnn} represents the entire time series information as a hypervariate graph and employs Fourier Graph Neural Network for global dependency modeling. On the other hand, variable-independent methods focus solely on modeling temporal dependencies under the assumption of variable independence~\cite{Zeng2023Dlinear, nie2023a}. These approaches are known for their simplicity and efficiency, often capable of mitigating model overfitting and achieving remarkable outcomes~\cite{Zeng2023Dlinear, nie2023a}. Nevertheless, this assumption may oversimplify the problem and potentially lead to ill-posed scenarios~\cite{zhang2023crossformer, liu2024itransformer}.


\paragraph{State Space Models}
Recently, some works~\cite{voelker2019legendre, gu2021combining} have combined SSMs with deep learning and demonstrated significant potential for capturing long-range dependencies. However, the prohibitive computation and memory costs of state representations often limit their practical applications~\cite{gu2021efficiently}. Several efficient variants of SSMs, such as S4~\cite{gu2021efficiently}, H3~\cite{fu2022h3}, Gated State Space~\cite{mehta2022long}, and RWKV~\cite{peng2023rwkv}, have been proposed to improve scalability and performance. Further, Mamba~\cite{gu2023mamba} introduces a data-dependent selection mechanism based on S4, enabling efficient filtering of relevant inputs and capturing long-range context that scales linearly with sequence length. This design allows Mamba to achieve linear-time efficiency while outperforming Transformer models on benchmark tasks.

Mamba has been successfully extended beyond sequential modeling to domains such as images~\cite{liu2024vmamba, al2026conmamba}, point clouds~\cite{liang2024pointmamba}, tables~\cite{ahamed2024mambatab}, and graphs~\cite{behrouz2024graph}, demonstrating strong capability in capturing long-range dependencies with linear computational complexity. To improve robustness to scan-order sensitivity, existing variants often adopt bidirectional~\cite{li2024videomamba, liang2024bi}, multi-directional~\cite{li2024mamba, liu2024vmamba}, or adaptive scanning strategies~\cite{huang2024localmamba} tailored to specific structural priors.

In LTSF, recent methods such as TimeMachine~\cite{ahamed2024timemachine} and Bi-Mamba4TS~\cite{liang2024bi} enhance temporal and inter-variable modeling through improved sequence representations. However, these approaches primarily focus on improving the SSM backbone while implicitly assuming a fixed or uninformative variable order~\cite{wang2025mamba}. This assumption is non-trivial in multivariate settings, since adjacent variables in a flattened scan may not correspond to meaningful statistical or causal relationships, potentially introducing undesirable locality bias. Motivated by this observation, we propose VAST, which explicitly estimates inter-variable dependencies and constructs a dependency-aware scan order for structured sequence modeling. To our knowledge, MambaTS is the first framework to explicitly investigate the role of variable scan order in SSM-based multivariate time series forecasting.

\section{Preliminaries}
\label{preliminaries}

\paragraph{State Space Models} SSMs are typically formulated as linear time-invariant (LTI) systems that map continuous inputs $x(t)$ to outputs $y(t)$ via a latent state $h(t)$. The state evolution can be described by ordinary differential equations as follows:
\begin{equation}
	\begin{aligned}
		h^\prime(t) & =\boldsymbol{A}h(t)+\boldsymbol{B}x(t) \\
		y(t) & =\boldsymbol{C}h(t)+\boldsymbol{D}x(t)
	\end{aligned}
\end{equation}
Here, $h^\prime(t)=\frac{dh(t)}{dt}$, and $\boldsymbol{A}, \boldsymbol{B}, \boldsymbol{C}$, and $\boldsymbol{D}$ are parameters of the time-independent SSMs.

\paragraph{Discretization} Finding analytical solutions for SSMs is highly challenging due to their continuous nature. Discretization is typically employed to facilitate analysis and solution in the discrete domain, which involves approximating the continuous-time state space model into a discrete-time representation. This is done by sampling the input signals at fixed time intervals to obtain their discrete-time counterparts. The resulting discrete-time state space model can be represented as:
\begin{equation}
	\begin{aligned}
		h_k & =\overline{\boldsymbol{A}} h_{k-1}+\overline{\boldsymbol{B}} x_k \\
		y_k & =\overline{\boldsymbol{C}} h_k+\overline{\boldsymbol{D}} x_k
	\end{aligned}
\end{equation}
Here, $h_k$ represents the state vector at time instant $k$, and $x_k$ represents the input vector at time instant $k$. The matrices $\overline{A}$ and $\overline{B}$ are derived from the continuous-time matrices $A$ and $B$ using appropriate discretization techniques such as the Euler or ZOH (Zero-Order Hold) method. In this case, $\overline{\boldsymbol{A}}=\exp (\Delta \boldsymbol{A}), \overline{\boldsymbol{B}}=(\Delta \boldsymbol{A})^{-1}(\exp (\Delta \boldsymbol{A})-\boldsymbol{I}) \cdot \Delta \boldsymbol{B}$.

\paragraph{Selective Scan Mechanism}
Mamba~\cite{gu2023mamba} further introduces selective SSMs by allowing the parameters to influence the interactions along the sequence in a context-dependent manner. This selective mechanism enables Mamba to filter out irrelevant noise, while selectively propagating or forgetting information relevant to the current input. This differs from previous SSMs methods with static parameters, but it does break the LTI characteristics. Therefore, Mamba takes a hardware optimization approach and implements parallel scan training to address this challenge.


\section{Motivation}
\label{motivation}

For the multivariate time series forecasting problem, we consider a look-back window of length \(L\) for \(K\) variables, denoted as \((\mathbf{x}_1, \mathbf{x}_2, \cdots, \mathbf{x}_K)\), where each \(\mathbf{x}_i \in \mathbb{R}^L\) represents the values of variable \(i\) over the past \(L\) time steps. Our objective is to forecast the values of the future \(T\) time steps for each variable, denoted as \((\hat{\mathbf{y}}_1, \hat{\mathbf{y}}_2, \cdots, \hat{\mathbf{y}}_K)\), where each \(\hat{\mathbf{y}}_i \in \mathbb{R}^T\).

Recent studies~\cite{Zeng2023Dlinear, nie2023a} have introduced the variable independence assumption, which reformulates the multivariate LTSF into multiple univariate LTSF. However, this assumption can be overly simplistic and may lead to ill-posed scenarios~\cite{zhang2023crossformer}. In practical applications, multivariate time series often exhibit causal relationships; neglecting these dependencies can result in suboptimal predictive performance. Therefore, modeling variable dependencies remains a dominant approach, particularly exemplified by Transformer-based methods that operate with quadratic complexity in pairwise computations~\cite{liu2024itransformer}. Some researches, such as~\cite{zhang2023crossformer}, seek to mitigate this complexity through a router mechanism that aggregates information from all variables prior to redistribution. Nonetheless, these methods still struggle with efficiency and scalability.

Given these challenges, we pose the question: \emph{Is it possible to achieve global dependency modeling with linear complexity without sacrificing the integrity of variable dependencies?} The answer lies in leveraging the structural properties of causal relationships among variables. By representing these relationships through causal graphs, we can effectively capture the fundamental dependencies that drive multivariate interactions. The following theorem formalizes this approach:

\begin{proposition}
	\label{prop:linear_scan}
	For a multivariate time series dataset, if a causal graph \( G = (V, E) \) exists that represents the relationships among the variables \( \mathbf{V} = \{\mathbf{V}_1, \mathbf{V}_2, \ldots, \mathbf{V}_K\} \), then global dependency modeling can be achieved through a single linear scan.
\end{proposition}

\begin{proof}
	A causal graph \( G \) is constructed, where \( V \) represents the variables and \( E \) indicates the causal relationships. Since \( G \) is a directed acyclic graph (DAG), there exists at least one topological order \( \sigma \) such that for any directed edge \( (\mathbf{V}_i, \mathbf{V}_j) \), \( \mathbf{V}_i \) precedes \( \mathbf{V}_j \) in this order. Following this topological ordering during a linear scan, each variable \( \mathbf{V}_{\sigma(k)} \) can be conditioned on its parents \( \text{Pa}(\mathbf{V}_{\sigma(k)}) \), allowing us to represent dependencies as \( P(\mathbf{V}_{\sigma(k)} | \text{Pa}(\mathbf{V}_{\sigma(k)}) ) \). Consequently, the joint distribution of all variables is obtained as \( P(\mathbf{V}) = \prod_{i=1}^{K} P(\mathbf{V}_{\sigma(i)} | \text{Pa}(\mathbf{V}_{\sigma(i}) ) \). This process demonstrates that a single linear scan effectively models global dependencies in multivariate time series by leveraging the structural information encoded in the causal graph.
\end{proof}

Despite Proposition~\ref{prop:linear_scan} demonstrating the potential for global dependency modeling with known causal relationships, real-world datasets often lack explicit dependency among variables. As a result, the underlying causal structure can be obscured by data complexity and interactions. To address this, it is essential to adopt methods for inferring causal relationships without a predefined graph. A promising approach is the random walk~\cite{Przymus2017ImprovingMT,kim2024revisiting}, which employs stochastic processes to explore the graph's structure and estimate causal links based on empirical transition observations.

\begin{definition}[\textbf{Random Walk Without Return}]
	\label{def:rwwr}
	A {random walk without return} is a stochastic process defined on a graph where a walker starts at a node and moves to an adjacent node uniformly at random, with the restriction that it cannot return to previously visited nodes until all nodes have been visited.
\end{definition}

\begin{proposition}
	\label{prop:rwwr}
	Given a causal graph \( G=(V, E) \) with unknown relationships among nodes \( \mathbf{V} = \{\mathbf{V}_1, \mathbf{V}_2, \ldots, \mathbf{V}_K\} \), if the total cost of a random walk without return is known, then the causal relationships can be estimated through infinite random walks without return.
\end{proposition}

\begin{figure*}[!ht]
	\centering
	\includegraphics[width=1\linewidth]{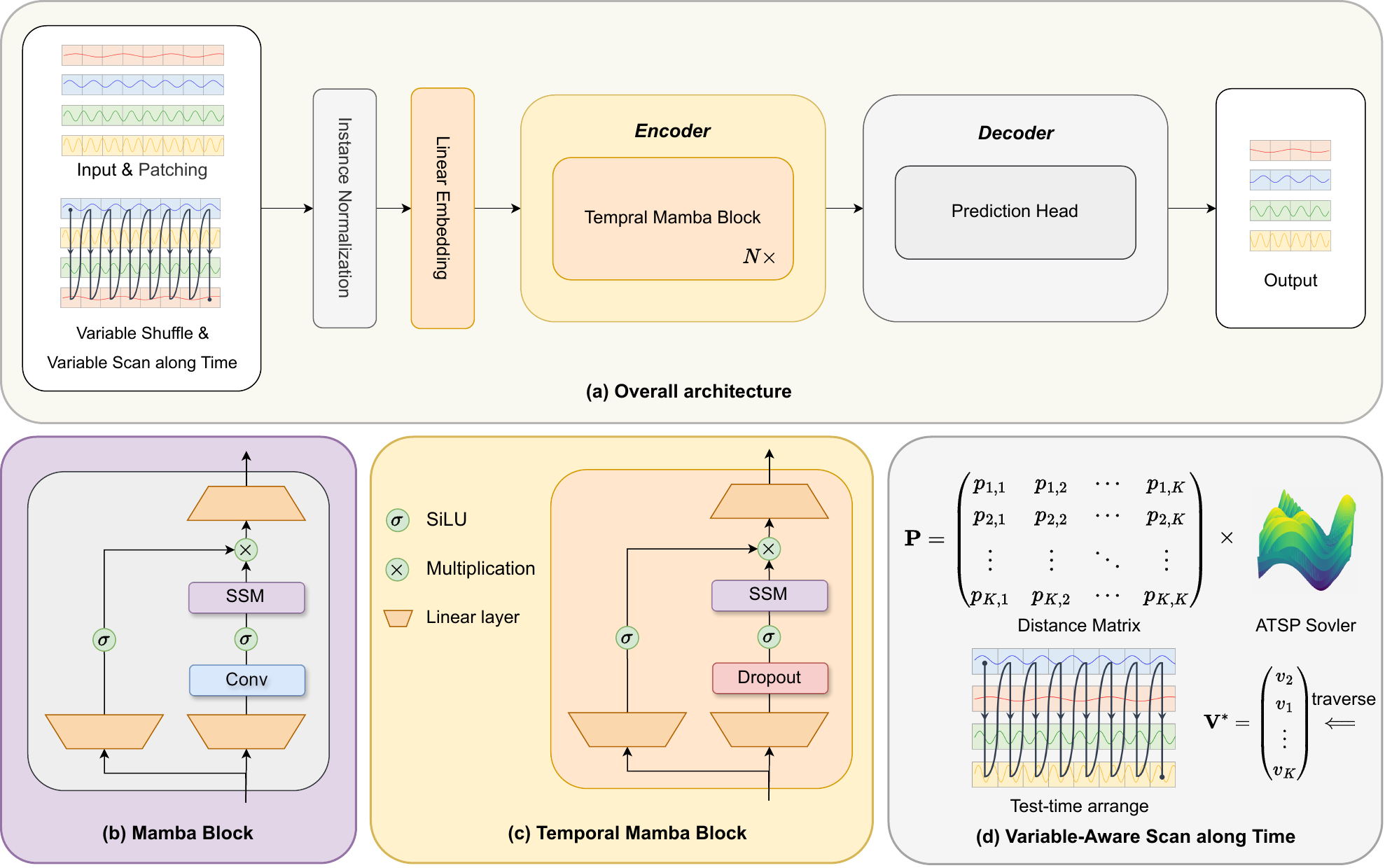}
	\caption{The overall architecture of MambaTS.}
	\label{fig:arch}
\end{figure*}

\begin{proof}
	In a random walk without return, the total cost \( C \) is shared across \( K-1 \) transitions. Assuming that each transition contributes evenly to \( C \), we focus on whether \( C \) accurately reflects the true transition cost. Transitions are classified into three types: positive transitions (PT), where \( C_{i,j}^{\text{PT}} > 0 \) if \( \mathbf{V}_i \to \mathbf{V}_j \) is a causal relationship; negative transitions (NT), where \( C_{i,j}^{\text{NT}} = -C_{j,i}^{\text{PT}}\) if \( \mathbf{V}_j \to \mathbf{V}_i \) is causal; and independent node transitions (IN), where \( 0 < C_{i,j}^{\text{IN}} < C_{i,j}^{\text{PT}} \) if no causal relationship exists. Due to the symmetry of the graph, \( \#\text{PT} = \#\text{NT} \), and thus at least \( \frac{\#\text{PT} + \#\text{IN}}{\#\text{PT} + \#\text{NT} + \#\text{IN}} \geq \frac{1}{2} \) of the transitions contribute to the cost update, with equality if and only if \( \#\text{IN} = 0 \). As \( N \to \infty \), the expected cost for each transition converges to a positive value, \( \mathbb{E}[C^{(n)}_{i,j}] > 0 \), and the average cost after \( N \) random walks is \( p_{i,j} = \frac{1}{N} \sum_{n=1}^{N} C_{i,j}^{(n)} \). Finally, the strength of the causal relationship between \( \mathbf{V}_i \) and \( \mathbf{V}_j \) is estimated as \( \hat{R}_{i,j} = \frac{p_{i,j}}{\sum_{k \in V} p_{i,k}} \), allowing causal relationships to be accurately estimated as \( N \to \infty \). 
\end{proof}

Through Proposition~\ref{prop:linear_scan} and~\ref{prop:rwwr}, we establish that for any multivariate dataset that can be represented by a causal graph, global dependency modeling can be efficiently accomplished with linear complexity. In the next section, we will present a detailed instantiation of this framework, termed MambaTS, which integrates these theoretical underpinnings into a practical methodology for effective LTSF.

\section{Methodology}
\label{model_architecture}

\subsection{Overall Architecture}

For clarity, We assume that the order of the $K$ variables satisfies the linear scanning condition, and we will present the estimation of variable order relationship in Section~\ref{subs:vast}. The architecture of MambaTS is illustrated in Figure~\ref{fig:arch}. It primarily consists of an embedding layer, an instance normalization layer, $N\times$ Temporal Mamba blocks, and a prediction head.

\paragraph{Patching and Embedding} As shown in Figure~\ref{fig:arch}~(a), for the input \(\mathbf{x}=(\mathbf{x}_1, \mathbf{x}_2, \cdots, \mathbf{x}_K)\in\mathbb{R}^{K\times L}\), we adopt the segmentation approach used in PatchTST, dividing each variable into patches every \(s\) time steps. This process yields \(M=\lceil\frac{L}{s}\rceil\) patches for each variable, where \(\mathbf{x}_k=(\mathbf{x}_k^{(1)}, \mathbf{x}_k^{(2)}, \cdots, \mathbf{x}_k^{(M)})\) for \(k = 1, 2, \ldots, K\). These patches are then embedded into \(D\)-dimensional tokens via a linear mapping defined as \(\mathbf{z}_k^{(j)} = \mathbf{W} \cdot \mathbf{x}_k^{(j)}\), where \(\mathbf{W} \in \mathbb{R}^{D \times s}\) is the weight matrix.

\paragraph{Variable Scan along Time} By embedding \(K\) variables, we obtain \(K \times M\) tokens, which are interleaved into \(\mathbf{z}=(\mathbf{z}_1^{(1)}, \mathbf{z}_2^{(1)}, \ldots, \mathbf{z}_1^{(2)}, \mathbf{z}_2^{(2)}, \ldots, \mathbf{z}_{K-1}^{(M)}, \mathbf{z}_K^{(M)})\), referred to as Variable Scan along Time (VST). Unlike PatchTST~\cite{nie2023a} (temporal-level) or iTransformer~\cite{liu2024itransformer} (variable-level), VST adopts a fine-grained interleaved structure that better preserves retrospective information. The resulting tokens are then fed into the encoder to model global dependencies across time and variables.

\paragraph{Encoder} The encoder consists of \(N\) stacked Temporal Mamba Block (TMB), modified from Mamba Block~\cite{gu2023mamba}. The architecture of the Mamba Block is illustrated in Figure~\ref{fig:arch}~(b) and features two branches: the right branch focuses on sequence modeling, while the left contains a gated nonlinear layer. The computation process of the Mamba Block is defined as follows:
\begin{equation}
	\mathbf{h} = \text{SSM}(\text{Conv}(\text{Linear}(\mathbf{z}))) + \sigma(\text{Linear}(\mathbf{z})).
\end{equation}
Here, the causal convolution $\text{Conv}$, acting as a shift-SSM, is inserted before the main SSM layer to enhance connections between adjacent tokens. However, given that the \(K\) variables may exhibit independence, the TMB (see Figure~\ref{fig:arch}~(c)) removes this component. Additionally, the TMB introduce a dropout mechanism for selective parameters to mitigate overfitting, as detailed below:
\begin{equation}
	\mathbf{h} = \text{SSM}(\text{Dropout}(\text{Linear}(\mathbf{z}))) + \sigma(\text{Linear}(\mathbf{z})).
\end{equation}

\paragraph{Prediction Head} During the decoding phase, akin to PatchTST~\cite{nie2023a}, we predict independently for each variable. This process involves a simple linear head that transforms the encoded result ${\mathbf{h}}=(\mathbf{h}_1, \mathbf{h}_2, \cdots, \mathbf{h}_K) \in \mathbb{R}^{K \times (MD)}$ to $\hat{\mathbf{y}} \in\mathbb{R}^{K \times T}$. This setup ensures the model to generate forecast results effectively.

\paragraph{Instance Normalization} To mitigate distribution shift between training and test data, following RevIN~\cite{kim2021reversible}, we standardize each input channel to zero mean and unit standard deviation, and retain these statistics for the de-normalization of model predictions.

\paragraph{Loss Function} We adopt mean squared error (MSE) loss as the primary loss function, given by:
\begin{equation}
	\mathcal{L} = \frac{1}{K} \sum_{i=1}^K\left\|\hat{\mathbf{y}}_{i}-\mathbf{y}_{i}\right\|_2^2 .
	\label{eq:loss}
\end{equation}

\subsection{Variable-Aware Scan along Time}
\label{subs:vast}

As emphasized in Proposition~\ref{prop:linear_scan}, the sequence of variable scanning is essential for accurately modeling global variable dependencies. However, the relationships among variables are not known a priori. Motivated by Proposition~\ref{prop:rwwr}, we propose Variable-Aware Scan along Time (VAST). The core idea of VAST is to estimate inter-variable dependencies among variables during the training phase through random walks without return, while during inference, it determines the optimal linear scanning order using the shortest paths that traverse all nodes. 

\paragraph{Training.} In the training phase, we introduce the variable permutation training (VPT) strategy. Specifically, for the input \(\mathbf{x}=(\mathbf{x}_1, \mathbf{x}_2, \cdots, \mathbf{x}_K)\in\mathbb{R}^{K\times L}\), we randomly shuffle the order of variables in each iteration to perform a random walk without return. We then revert the shuffled state after decoding to ensure the correct output sequence.

For any $K$ variables, we maintain a directed graph adjacency matrix $\boldsymbol{P}\in\R^{K\times K}$, where $p_{i,j}$ represents the cost from node $i$ to node $j$. Notably, with the introduction of VPT, we can explore various combinations of scan orders and evaluate their value using Eq.~\ref{eq:loss}. Following a permutation, a node index sequence $\mathbf{V}=\{v_1,v_2,\cdots,v_K\}$ is obtained, where $v_k$ denotes the new index in the shuffled sequence. Subsequently, $K-1$ transition tuples $\{(v_1,v_2), (v_2, v_3),\cdots(v_{K-1},v_{K}) \}$ are derived. For each sample, a training loss ${l}^{(t)}$ is generated in the $t$-th iteration of the network. Therefore, we update $\boldsymbol{P}$ with exponential moving average:
\begin{equation}
	p^{(t)}_{v_k,v_{k+1}}=\beta p^{(t-1)}_{v_k,v_{k+1}}+(1-\beta)l^{(t)},
	\label{eq:updata_p}
\end{equation}
where $\beta$ is a hyperparameter controlling the sliding average rate, determining the impact of new estimates on the global variable importance. To facilitate efficient training, we extend the above formula to a batch version. By a simple nondimensionalization operation to eliminate the influence of different sample batches, we define $\bar{\textbf{l}}^{(t)}={\textbf{l}}^{(t)}-\text{dev}({\textbf{l}}^{(t)})$, where $\text{dev}$ denotes the standard deviation and \({\textbf{l}}^{(t)}\) refers to the batch version of \(l^{(t)}\). Consequently, the Eq.~\ref{eq:updata_p} is modified to:
\begin{equation}
	\boldsymbol{P}^{(t)}=\beta \boldsymbol{P}^{(t-1)} + (1-\beta)\bar{\textbf{l}}^{(t)}.
\end{equation}

\paragraph{Inference} Throughout training, $\boldsymbol{P}$ are leveraged to determine the optimal variable scan order. This involves solving the asymmetric traveling salesman problem (ATSP), which seeks the shortest path visiting all nodes. Given the dense connectivity represented by $\boldsymbol{P}$, finding the optimal traversal path is NP-hard. Hence, we introduce a heuristic-based simulated annealing~\cite{dreo2006metaheuristics} algorithm for path decoding.

\paragraph{Discussion}
Through VAST, we enable efficient modeling of variable dependencies via a learned scan mechanism guided by optimization signals. It is worth noting that the DAG assumption provides an idealized theoretical starting point for Proposition~\ref{prop:linear_scan}, which helps characterize the linear scan property under a simplified setting. In practice, VAST does not assume access to an explicit causal graph. Instead, it approximates this idealized process through a loss-driven heuristic estimation of variable relationships during training. The effectiveness of VAST, as well as the quality of the learned variable dependency structure, is further validated empirically in the experimental section.

\section{Experiments}
\label{experiments}

\paragraph{Dataset} We conducted extensive experiments on eight public datasets, as shown in Table~\ref{tab:datesets}, including two ETT datasets~\cite{zhou2021informer}, Weather, Electricity, Traffic~\cite{wu2021autoformer}, Solar~\cite{lai2018modeling}, Covid-19~\cite{panagopoulos2021transfer}, and PEMS~\cite{liu2022scinet}, covering domains such as electricity, energy, transportation, weather, and health. For all datasets, we strictly follow the data preprocessing and train/validation/test splits protocols as in iTransformer~\cite{liu2024itransformer}, to ensure fair and consistent comparisons.

\begin{table*}[!ht]
	\centering
	\caption{Summary of Dataset Characteristics}
	\label{tab:datesets}
	\begin{adjustbox}{max width=0.84\linewidth}
		\begin{tabular}{@{}ccccccccc@{}}
			\toprule
			Datasets  & ETTh2  & ETTm2  & Weather  & Electricity  & Traffic & Solar   & Covid-19 & PEMS \\ \midrule
			Features & 7      & 7      & 21      & 321      & 862         & 137     & 948     & 358      \\
			Time steps    & 17,420  & 17,420  & 52,696  & 26,304  & 17,544       & 52,179   & 1,392   & 21,351   \\ 
			Frequency & 1 hour & 15 mins & 10 mins & 1 hour & 1 hour      & 10 mins & 1 day  & 5 mins  \\
			\bottomrule
		\end{tabular}
	\end{adjustbox}
\end{table*}

\begin{table*}[!t]
	\caption{Multivariate long-term series forecasting results. All models employ a look-back window length of \(L=96\) for the Covid-19 dataset and \(L=720\) for the remaining datasets. Red bold and blue underlined values denote the best and second-best results, respectively. 
	}
	\label{tab:main}
	\begin{adjustbox}{max width=\linewidth}
		\begin{tabular}{@{}cccccccccccccccccc@{}}
			\toprule
			\multicolumn{2}{c}{Models}                                                   & \multicolumn{2}{c}{MambaTS} & \multicolumn{2}{c}{iTransformer} & \multicolumn{2}{c}{FourierGNN} & \multicolumn{2}{c}{PatchTST} & \multicolumn{2}{c}{Dlinear} & \multicolumn{2}{c}{MICN} & \multicolumn{2}{c}{FEDformer} & \multicolumn{2}{c}{Autoformer} \\ \midrule
			\multicolumn{2}{c}{Metric}                                                   & MSE                                    & MAE                                    & MSE                                   & MAE                                   & MSE                                   & MAE                                  & MSE                                    & MAE                              & MSE                                         & MAE                                & MSE                                       & MAE                                & MSE                                    & MAE                                    & MSE                                   & MAE                                   \\ 
			
			\midrule
			\multicolumn{1}{c|}{\multirow{4}{*}{ETTh2}} & \multicolumn{1}{c|}{96} & \textcolor{red}{\textbf{0.281}} & \textcolor{red}{\textbf{0.347}} & 0.312 & 0.363 & 0.454 & 0.481 & \textcolor{blue}{\ul 0.283} & \textcolor{red}{\textbf{0.347}} & 0.306 & 0.370 & 0.289 & \textcolor{blue}{\ul 0.354} & 0.332 & 0.374 & 0.332 & 0.368 \\
			\multicolumn{1}{c|}{} & \multicolumn{1}{c|}{192} & \textcolor{red}{\textbf{0.352}} & \textcolor{blue}{\ul 0.397} & 0.384 & 0.408 & 0.560 & 0.541 & \textcolor{blue}{\ul 0.354} & \textcolor{red}{\textbf{0.391}} & 0.411 & 0.437 & 0.408 & 0.444 & 0.407 & 0.446 & 0.426 & 0.434 \\
			\multicolumn{1}{c|}{} & \multicolumn{1}{c|}{336} & \textcolor{red}{\textbf{0.372}} & \textcolor{blue}{\ul 0.416} & 0.431 & 0.444 & 0.608 & 0.568 & \textcolor{blue}{\ul 0.376} & \textcolor{red}{\textbf{0.411}} & 0.542 & 0.514 & 0.547 & 0.516 & 0.400 & 0.447 & 0.477 & 0.479 \\
			\multicolumn{1}{c|}{} & \multicolumn{1}{c|}{720} & \textcolor{blue}{\ul 0.404} & \textcolor{blue}{\ul 0.444} & 0.432 & 0.463 & 0.820 & 0.648 & \textcolor{red}{\textbf{0.402}} & \textcolor{red}{\textbf{0.440}} & 0.900 & 0.671 & 0.834 & 0.688 & 0.412 & 0.469 & 0.453 & 0.490 \\
			
			\midrule
			
			\multicolumn{1}{c|}{\multirow{4}{*}{ETTm2}} & \multicolumn{1}{c|}{96} & \textcolor{blue}{\ul 0.166} & 0.260 & 0.181 & 0.275 & 0.229 & 0.327 & 0.168 & \textcolor{blue}{\ul 0.259} & \textcolor{red}{\textbf{0.163}} & \textcolor{red}{\textbf{0.258}} & 0.177 & 0.274 & 0.180 & 0.271 & 0.205 & 0.293 \\
			\multicolumn{1}{c|}{} & \multicolumn{1}{c|}{192} & \textcolor{blue}{\ul 0.228} & \textcolor{blue}{\ul 0.306} & 0.243 & 0.315 & 0.308 & 0.384 & 0.237 & 0.309 & \textcolor{red}{\textbf{0.222}} & \textcolor{red}{\textbf{0.304}} & 0.236 & 0.310 & 0.252 & 0.318 & 0.278 & 0.336 \\
			\multicolumn{1}{c|}{} & \multicolumn{1}{c|}{336} & \textcolor{blue}{\ul 0.276} & \textcolor{red}{\textbf{0.335}} & 0.297 & 0.352 & 0.362 & 0.413 & 0.279 & \textcolor{blue}{\ul 0.336} & \textcolor{red}{\textbf{0.274}} & \textcolor{blue}{\ul 0.336} & 0.299 & 0.350 & 0.324 & 0.364 & 0.343 & 0.379 \\
			\multicolumn{1}{c|}{} & \multicolumn{1}{c|}{720} & \textcolor{red}{\textbf{0.355}} & \textcolor{blue}{\ul 0.391} & 0.381 & 0.404 & 0.482 & 0.487 & \textcolor{blue}{\ul 0.363} & \textcolor{red}{\textbf{0.390}} & 0.407 & 0.432 & 0.421 & 0.434 & 0.410 & 0.420 & 0.414 & 0.419 \\

			\midrule
			\multicolumn{1}{c|}{\multirow{4}{*}{Weather}} & \multicolumn{1}{c|}{96} & \textcolor{red}{\textbf{0.145}} & \textcolor{red}{\textbf{0.195}} & 0.180 & 0.232 & 0.162 & 0.232 & \textcolor{blue}{\ul 0.149} & \textcolor{blue}{\ul 0.198} & 0.168 & 0.227 & 0.167 & 0.231 & 0.238 & 0.314 & 0.249 & 0.329 \\
			\multicolumn{1}{c|}{} & \multicolumn{1}{c|}{192} & \textcolor{red}{\textbf{0.192}} & \textcolor{red}{\textbf{0.241}} & 0.228 & 0.270 & 0.207 & 0.276 & \textcolor{blue}{\ul 0.194} & \textcolor{red}{\textbf{0.241}} & 0.212 & \textcolor{blue}{\ul 0.267} & 0.212 & 0.271 & 0.275 & 0.329 & 0.325 & 0.370 \\
			\multicolumn{1}{c|}{} & \multicolumn{1}{c|}{336} & \textcolor{red}{\textbf{0.245}} & \textcolor{red}{\textbf{0.280}} & 0.291 & 0.316 & 0.261 & 0.318 & \textcolor{red}{\textbf{0.245}} & \textcolor{blue}{\ul 0.282} & \textcolor{blue}{\ul 0.256} & 0.305 & 0.275 & 0.337 & 0.339 & 0.377 & 0.351 & 0.391 \\
			\multicolumn{1}{c|}{} & \multicolumn{1}{c|}{720} & \textcolor{blue}{\ul 0.313} & \textcolor{red}{\textbf{0.329}} & 0.354 & 0.359 & 0.336 & 0.366 & 0.314 & \textcolor{blue}{\ul 0.334} & 0.315 & 0.355 & \textcolor{red}{\textbf{0.312}} & 0.349 & 0.389 & 0.409 & 0.415 & 0.426 \\

			\midrule
			\multicolumn{1}{c|}{\multirow{4}{*}{Electricity}} & \multicolumn{1}{c|}{96} & \textcolor{red}{\textbf{0.128}} & \textcolor{red}{\textbf{0.223}} & \textcolor{blue}{\ul 0.130} & \textcolor{red}{\textbf{0.223}} & 0.133 & \textcolor{blue}{\ul 0.229} & 0.133 & 0.230 & 0.151 & 0.260 & 0.166 & 0.274 & 0.186 & 0.302 & 0.196 & 0.313 \\
			\multicolumn{1}{c|}{} & \multicolumn{1}{c|}{192} & \textcolor{red}{\textbf{0.146}} & \textcolor{red}{\textbf{0.239}} & \textcolor{blue}{\ul 0.147} & \textcolor{blue}{\ul 0.240} & 0.155 & 0.251 & \textcolor{blue}{\ul 0.147} & 0.244 & 0.165 & 0.276 & 0.182 & 0.289 & 0.197 & 0.311 & 0.211 & 0.324 \\
			\multicolumn{1}{c|}{} & \multicolumn{1}{c|}{336} & \textcolor{red}{\textbf{0.161}} & \textcolor{blue}{\ul 0.258} & 0.164 & \textcolor{red}{\textbf{0.257}} & 0.167 & 0.264 & \textcolor{blue}{\ul 0.162} & 0.261 & 0.183 & 0.291 & 0.201 & 0.308 & 0.213 & 0.328 & 0.214 & 0.327 \\
			\multicolumn{1}{c|}{} & \multicolumn{1}{c|}{720} & \textcolor{red}{\textbf{0.187}} & \textcolor{red}{\textbf{0.283}} & 0.203 & 0.292 & \textcolor{blue}{\ul 0.194} & \textcolor{blue}{\ul 0.288} & 0.196 & 0.294 & 0.201 & 0.312 & 0.235 & 0.339 & 0.233 & 0.344 & 0.236 & 0.342 \\
			
			\midrule
			\multicolumn{1}{c|}{\multirow{4}{*}{Traffic}} & \multicolumn{1}{c|}{96} & \textcolor{red}{\textbf{0.347}} & \textcolor{red}{\textbf{0.248}} & \textcolor{blue}{\ul 0.349} & 0.255 & 0.494 & 0.303 & 0.367 & \textcolor{blue}{\ul 0.253} & 0.385 & 0.269 & 0.445 & 0.295 & 0.576 & 0.359 & 0.597 & 0.371 \\
			\multicolumn{1}{c|}{} & \multicolumn{1}{c|}{192} & \textcolor{red}{\textbf{0.358}} & \textcolor{red}{\textbf{0.255}} & \textcolor{blue}{\ul 0.359} & 0.263 & 0.513 & 0.310 & 0.382 & \textcolor{blue}{\ul 0.259} & 0.395 & 0.273 & 0.461 & 0.302 & 0.610 & 0.380 & 0.607 & 0.382 \\
			\multicolumn{1}{c|}{} & \multicolumn{1}{c|}{336} & \textcolor{red}{\textbf{0.372}} & \textcolor{red}{\textbf{0.262}} & \textcolor{blue}{\ul 0.379} & 0.272 & 0.534 & 0.320 & 0.396 & \textcolor{blue}{\ul 0.267} & 0.409 & 0.281 & 0.483 & 0.307 & 0.608 & 0.375 & 0.623 & 0.387 \\
			\multicolumn{1}{c|}{} & \multicolumn{1}{c|}{720} & \textcolor{red}{\textbf{0.416}} & \textcolor{red}{\textbf{0.284}} & \textcolor{blue}{\ul 0.417} & 0.291 & 0.597 & 0.346 & 0.433 & \textcolor{blue}{\ul 0.287} & 0.449 & 0.305 & 0.527 & 0.310 & 0.621 & 0.375 & 0.639 & 0.395 \\
			
			\midrule
			\multicolumn{1}{c|}{\multirow{4}{*}{Solar}} & \multicolumn{1}{c|}{96} & \textcolor{red}{\textbf{0.165}} & \textcolor{red}{\textbf{0.231}} & \textcolor{blue}{\ul 0.170} & 0.246 & 0.183 & \textcolor{blue}{\ul 0.232} & 0.185 & 0.246 & 0.191 & 0.257 & 0.190 & 0.243 & 0.214 & 0.311 & 0.316 & 0.369 \\
			\multicolumn{1}{c|}{} & \multicolumn{1}{c|}{192} & \textcolor{red}{\textbf{0.178}} & \textcolor{red}{\textbf{0.240}} & \textcolor{blue}{\ul 0.195} & 0.263 & 0.198 & 0.256 & 0.201 & 0.262 & 0.211 & 0.273 & 0.205 & \textcolor{blue}{\ul 0.247} & 0.281 & 0.364 & 0.418 & 0.437 \\
			\multicolumn{1}{c|}{} & \multicolumn{1}{c|}{336} & \textcolor{red}{\textbf{0.192}} & \textcolor{blue}{\ul 0.252} & 0.217 & 0.282 & \textcolor{blue}{\ul 0.205} & 0.261 & 0.209 & 0.266 & 0.228 & 0.285 & 0.219 & \textcolor{red}{\textbf{0.250}} & 0.294 & 0.378 & 0.438 & 0.467 \\
			\multicolumn{1}{c|}{} & \multicolumn{1}{c|}{720} & \textcolor{red}{\textbf{0.199}} & \textcolor{red}{\textbf{0.258}} & 0.208 & 0.276 & \textcolor{blue}{\ul 0.202} & 0.265 & 0.226 & 0.283 & 0.236 & 0.294 & 0.227 & \textcolor{blue}{\ul 0.263} & 0.315 & 0.406 & 0.618 & 0.550 \\
			
			\midrule
			\multicolumn{1}{c|}{\multirow{4}{*}{Covid-19}} & \multicolumn{1}{c|}{12} & \textcolor{red}{\textbf{0.784}} & \textcolor{red}{\textbf{0.037}} & \textcolor{blue}{\ul 0.998} & \textcolor{blue}{\ul 0.046} & 3.584 & 0.075 & 1.236 & 0.054 & 2.643 & 0.087 & 6.505 & 0.110 & 7.607 & 0.316 & 7.695 & 0.406 \\
			\multicolumn{1}{c|}{} & \multicolumn{1}{c|}{24} & \textcolor{red}{\textbf{1.075}} & \textcolor{red}{\textbf{0.046}} & \textcolor{blue}{\ul 1.488} & \textcolor{blue}{\ul 0.060} & 2.532 & 0.079 & 1.584 & 0.064 & 3.678 & 0.100 & 23.587 & 0.155 & 8.162 & 0.312 & 8.253 & 0.410 \\
			\multicolumn{1}{c|}{} & \multicolumn{1}{c|}{48} & \textcolor{red}{\textbf{1.776}} & \textcolor{red}{\textbf{0.064}} & \textcolor{blue}{\ul 2.505} & \textcolor{blue}{\ul 0.082} & 12.922 & 0.140 & 2.639 & 0.089 & 5.836 & 0.131 & 33.467 & 0.206 & 9.458 & 0.328 & 9.563 & 0.440 \\
			\multicolumn{1}{c|}{} & \multicolumn{1}{c|}{96} & \textcolor{red}{\textbf{4.155}} & \textcolor{red}{\textbf{0.112}} & \textcolor{blue}{\ul 6.435} & \textcolor{blue}{\ul 0.146} & 7.991 & 0.164 & 11.811 & 0.176 & 10.092 & 0.185 & 24.247 & 0.261 & 12.694 & 0.550 & 12.592 & 0.456 \\
			
			\midrule
			\multicolumn{1}{c|}{\multirow{4}{*}{PEMS}} & \multicolumn{1}{c|}{12} & \textcolor{red}{\textbf{0.059}} & \textcolor{red}{\textbf{0.161}} & 0.064 & 0.167 & 0.091 & 0.202 & \textcolor{blue}{\ul 0.063} & \textcolor{blue}{\ul 0.166} & 0.078 & 0.187 & 0.094 & 0.204 & 0.283 & 0.394 & 0.584 & 0.607 \\
			\multicolumn{1}{c|}{} & \multicolumn{1}{c|}{24} & \textcolor{red}{\textbf{0.075}} & \textcolor{red}{\textbf{0.179}} & 0.081 & 0.187 & 0.116 & 0.232 & \textcolor{blue}{\ul 0.080} & \textcolor{blue}{\ul 0.185} & 0.113 & 0.224 & 0.116 & 0.229 & 0.300 & 0.431 & 0.672 & 0.664 \\
			\multicolumn{1}{c|}{} & \multicolumn{1}{c|}{48} & \textcolor{red}{\textbf{0.102}} & \textcolor{red}{\textbf{0.206}} & 0.111 & 0.215 & 0.165 & 0.271 & \textcolor{blue}{\ul 0.109} & \textcolor{blue}{\ul 0.213} & 0.167 & 0.274 & 0.147 & 0.255 & 0.396 & 0.476 & 0.879 & 0.781 \\
			\multicolumn{1}{c|}{} & \multicolumn{1}{c|}{96} & \textcolor{red}{\textbf{0.134}} & \textcolor{red}{\textbf{0.230}} & \textcolor{blue}{\ul 0.142} & \textcolor{blue}{\ul 0.240} & 0.196 & 0.300 & 0.145 & 0.243 & 0.212 & 0.313 & 0.256 & 0.362 & 0.477 & 0.537 & 1.100 & 0.895 \\
			
			\midrule
			\multicolumn{1}{c|}{Average} & \multicolumn{1}{c|}{--} & \textcolor{red}{\textbf{0.448}} & \textcolor{red}{\textbf{0.250}} & \textcolor{blue}{\ul 0.578} & 0.266 & 1.128 & 0.313 & 0.752 & \textcolor{blue}{\ul 0.258} & 0.942 & 0.289 & 3.002 & 0.303 & 1.490 & 0.379 & 1.585 & 0.439 \\
			
			\bottomrule
			\\ 
		\end{tabular}
	\end{adjustbox}
\end{table*}

\paragraph{Baselines and Metrics} To demonstrate the effectiveness of MambaTS, we compared it against SOTA models of LTSF, including five popular Transformer-based methods: PatchTST~\cite{nie2023a}, iTransformer~\cite{liu2024itransformer}, FEDformer~\cite{zhou2022fedformer}, Autoformer~\cite{wu2021autoformer}, and three representative non-Transformer-based methods: DLinear~\cite{Zeng2023Dlinear}, MICN~\cite{wang2023micn}, and FourierGNN~\cite{yi2023fouriergnn}. Following PatchTST~\cite{nie2023a}, we primarily evaluate the models using Mean Squared Error (MSE) and Mean Absolute Error (MAE).


\paragraph{Implementation Details} Experiments were performed on an NVIDIA RTX 3090 Ti 24 GB GPU using the Adam optimizer~\cite{kingma2014adam} with betas of (0.9, 0.999). Training ran for 10 epochs with early stopping implemented using a patience of 3 to avoid overfitting. The best parameter selection for all comparison models was carefully tuned on the validation set.

\begin{table*}[!ht]
	\caption{Ablations on components. VST: Variable Scan along Time. TMB: Temporal Mamba Block. VAST: Variable-Aware Scan along Time. The average results of all predicted lengths are listed here.}
	\label{tab:ablation_main}
	\vspace{5pt}
	\centering
	\begin{small}
		\fontsize{10pt}{12pt}\selectfont
		\centering
		\renewcommand{\arraystretch}{1.1}
		\scalebox{0.78}{
			\begin{tabular}{ccc||cc|cc|cc|cc}	
				\toprule
				\multirow{2}{*}{VST} &\multirow{2}{*}{TMB} & \multirow{2}{*}{VAST} & \multicolumn{2}{c|}{ETTm2} & \multicolumn{2}{c|}{Traffic} & \multicolumn{2}{c|}{Electricity} & \multicolumn{2}{c}{Solar}\\
				\cline{4-11}
				& & & MSE & MAE & MSE & MAE  & MSE & MAE & MSE & MAE \\
				\cline{1-11}
				$\Circle$ & $\Circle$ & $\Circle$ & 0.285 & 0.342 & 0.400 & 0.273 & 0.167 & 0.260 & 0.192 & 0.261 \\
				$\CIRCLE$ & $\Circle$ & $\Circle$ & 0.284 & 0.341 & 0.383 & 0.268 & 0.164 & 0.260 & 0.197 & 0.266 \\
				$\Circle$ & $\CIRCLE$ & $\Circle$ & 0.264 & 0.329 & 0.389 & 0.268 & 0.160 & 0.255 & 0.191 & 0.253 \\
				$\CIRCLE$ & $\CIRCLE$ & $\Circle$ & 0.263 & 0.327 & 0.376 & 0.267 & 0.161 & 0.257 & 0.193 & 0.261 \\
				\cline{1-11}
				$\CIRCLE$ & $\CIRCLE$ & $\CIRCLE$ & \bf 0.262 & \bf 0.325 & \bf 0.373 & \bf 0.262 & \bf 0.155 & \bf 0.251 & \bf 0.184 & \bf 0.247 \\
				\bottomrule
		\end{tabular}}
	\end{small}
\end{table*} 

\begin{figure*}[!t]
	\centering
	\includegraphics[width=1\linewidth]{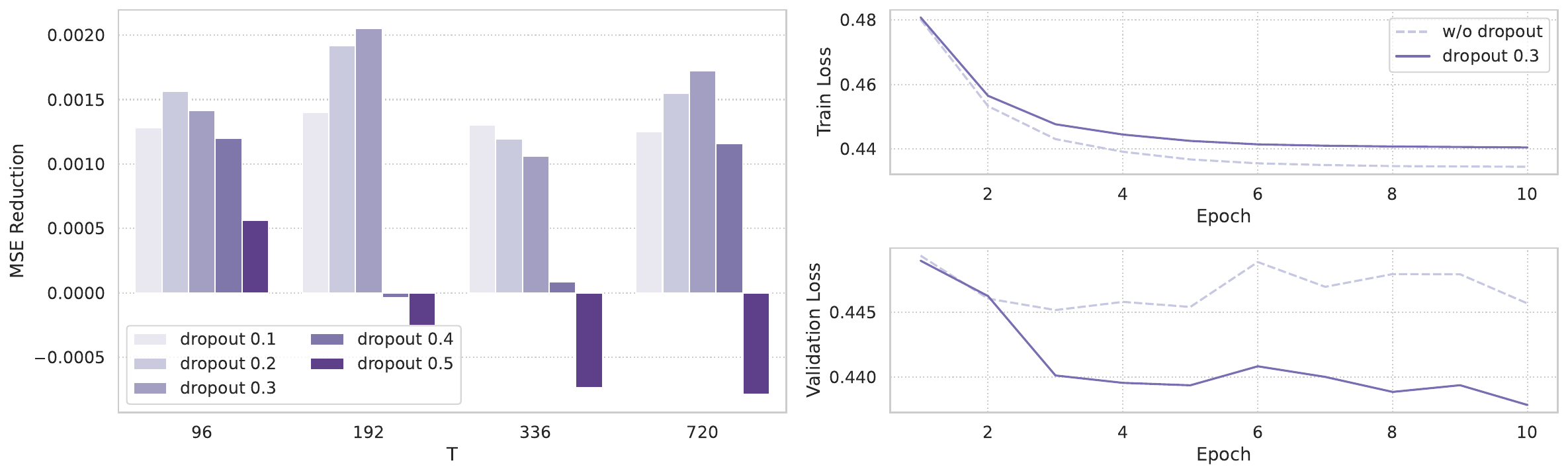}
	\caption{Dropout ablations of TMB. Left: TMB with varying dropout rates on Weather dataset. Right: Loss curves over training.}
	\label{fig:dropout}
\end{figure*}

\begin{figure*}[!h]
	\begin{minipage}[c]{.45\linewidth}
		\centering
		\fontsize{10pt}{12pt}\selectfont
		\centering
		\renewcommand{\arraystretch}{1.1}
		\scalebox{0.8}{
			\begin{tabular}{c|cc|cc|cc}	
				\toprule
				\multirow{2}{*}{\makecell{Scanning}} & \multicolumn{2}{c|}{ETTm2} & \multicolumn{2}{c|}{Traffic} & \multicolumn{2}{c}{Electricity} \\
				\cline{2-7}
				& MSE & MAE & MSE & MAE  & MSE & MAE \\
				\cline{1-7}
				W/o VPT & 0.263 & 0.327 & 0.376 & 0.267 & 0.161 & 0.257 \\
				Random (100x) & 0.262 & 0.326 & 0.374 & 0.265 & 0.158 & 0.256 \\
				VAST ($\text{GD.}$) & 0.260 & 0.322 & 0.376 & 0.267 & 0.161 & 0.257 \\
				VAST ($\text{LS.}$) & 0.261 & 0.325 & 0.375 & 0.265 & 0.157 & 0.254 \\
				VAST ($\text{LK.}$) & 0.262 & 0.325 & 0.374 & 0.264 & 0.156 & 0.252 \\
				\cline{1-7}
				VAST ($\text{SA.}$) & \bf 0.259 & \bf 0.321 & \bf 0.373 & \bf 0.262 & \bf 0.156 & \bf 0.251 \\
				\bottomrule
			\end{tabular}
		}
		\captionof{table}{Ablations on VAST.}
		\label{tab:vast}
	\end{minipage}
	\begin{minipage}[c]{.55\linewidth}
		\centering
		\includegraphics[width=0.93\linewidth]{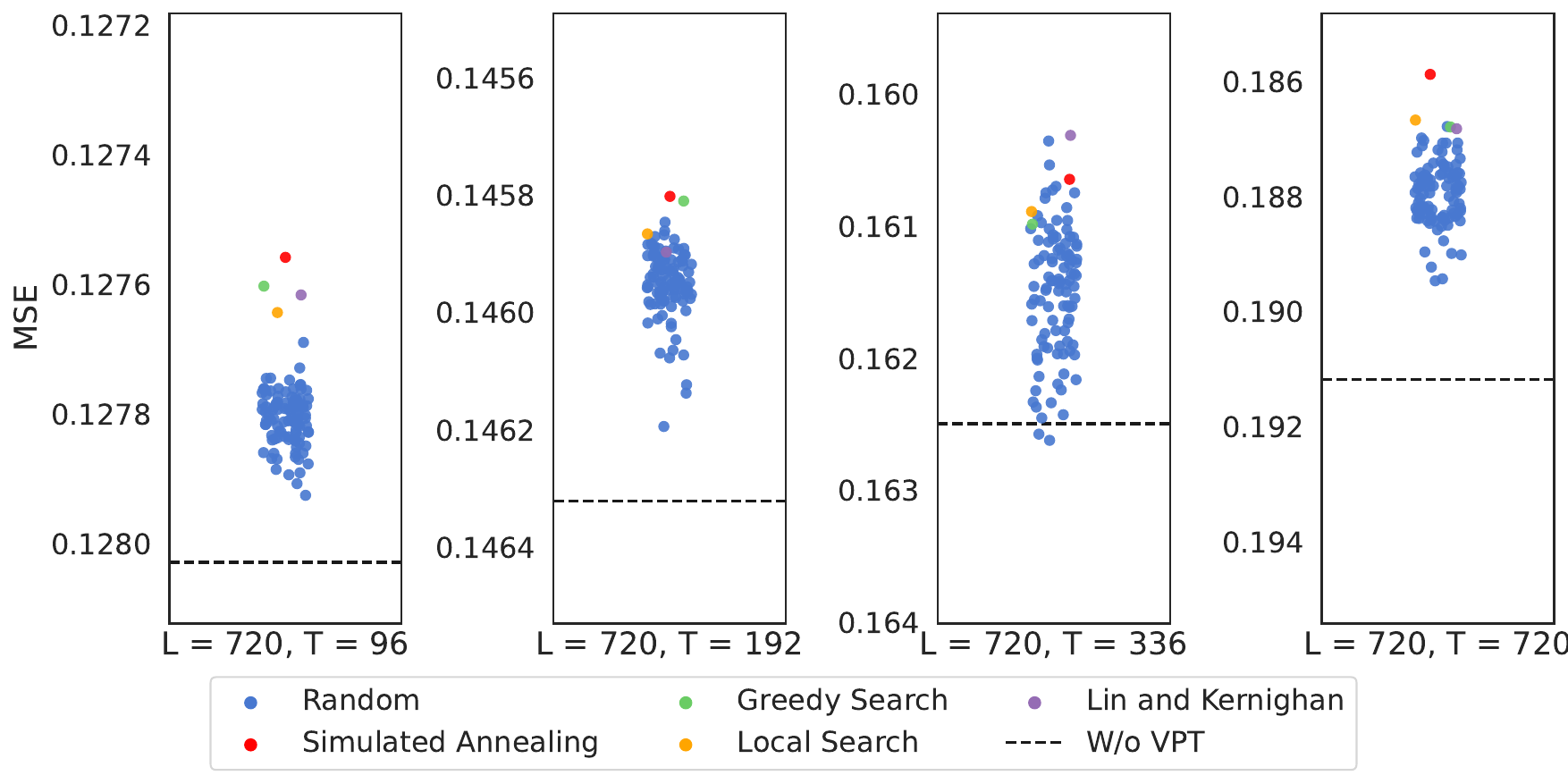}
		\captionof{figure}{Sampling results of VAST. Employing swarm plot to prevent overlapping points (better in color).}
		\label{fig:vast}
	\end{minipage}
\end{figure*} 

\subsection{Main results}
Table~\ref{tab:main} reports the results of multivariate long-term forecasting. Overall, MambaTS achieves competitive or state-of-the-art performance on most datasets and prediction horizons. A clear trend can be observed across different dataset regimes: methods such as DLinear and PatchTST, which rely on variable-independence assumptions, perform well on low-dimensional datasets (e.g., ETTh2/m2 with $K=7$ and Weather with $K=21$), but degrade on high-dimensional settings such as Traffic ($K=862$) and Covid-19 ($K=948$), where inter-variable dependencies are more complex. In contrast, iTransformer exhibits stronger performance on high-dimensional datasets but is less effective on low-dimensional ones, indicating a different inductive bias in modeling variable interactions. Compared with these methods, MambaTS achieves particularly strong performance on high-dimensional and complex datasets, while maintaining competitive results on simpler settings, suggesting that explicitly modeling variable dependencies via the proposed scan mechanism provides a more balanced representation under varying dataset complexities. Other baselines show competitive but less consistent behavior across datasets and forecasting horizons.

Further comparisons with recent Mamba-based forecasting methods are provided in Section~\ref{sec:additional_exp}.

\subsection{Ablation studies and analyses}

To validate the rationality and effectiveness of the proposed components, we conducted extensive ablation experiments as shown in Table~\ref{tab:ablation_main}, Table~\ref{tab:vast}, and Figure~\ref{fig:dropout}.

\paragraph{Component Ablation} Table~\ref{tab:ablation_main} reports the ablation results of key components. We first observe that incorporating VST consistently improves performance over the baseline Mamba-based PatchTST across most datasets, indicating the importance of explicitly modeling inter-variable interactions. Replacing the backbone with TMB further improves performance, demonstrating the effectiveness of the proposed temporal modeling design. When combining VST and TMB, the model achieves strong overall results, while slight variations appear on certain datasets, suggesting that jointly modeling variable-level dependencies and temporal dynamics introduces a more coupled optimization setting compared to each component alone. Finally, integrating VAST yields the most consistent performance across datasets (see Table~\ref{tab:ablation_main}), validating the effectiveness of the full design and the synergy between components.

\begin{figure*}[!t]
	\centering
	\includegraphics[width=1\linewidth]{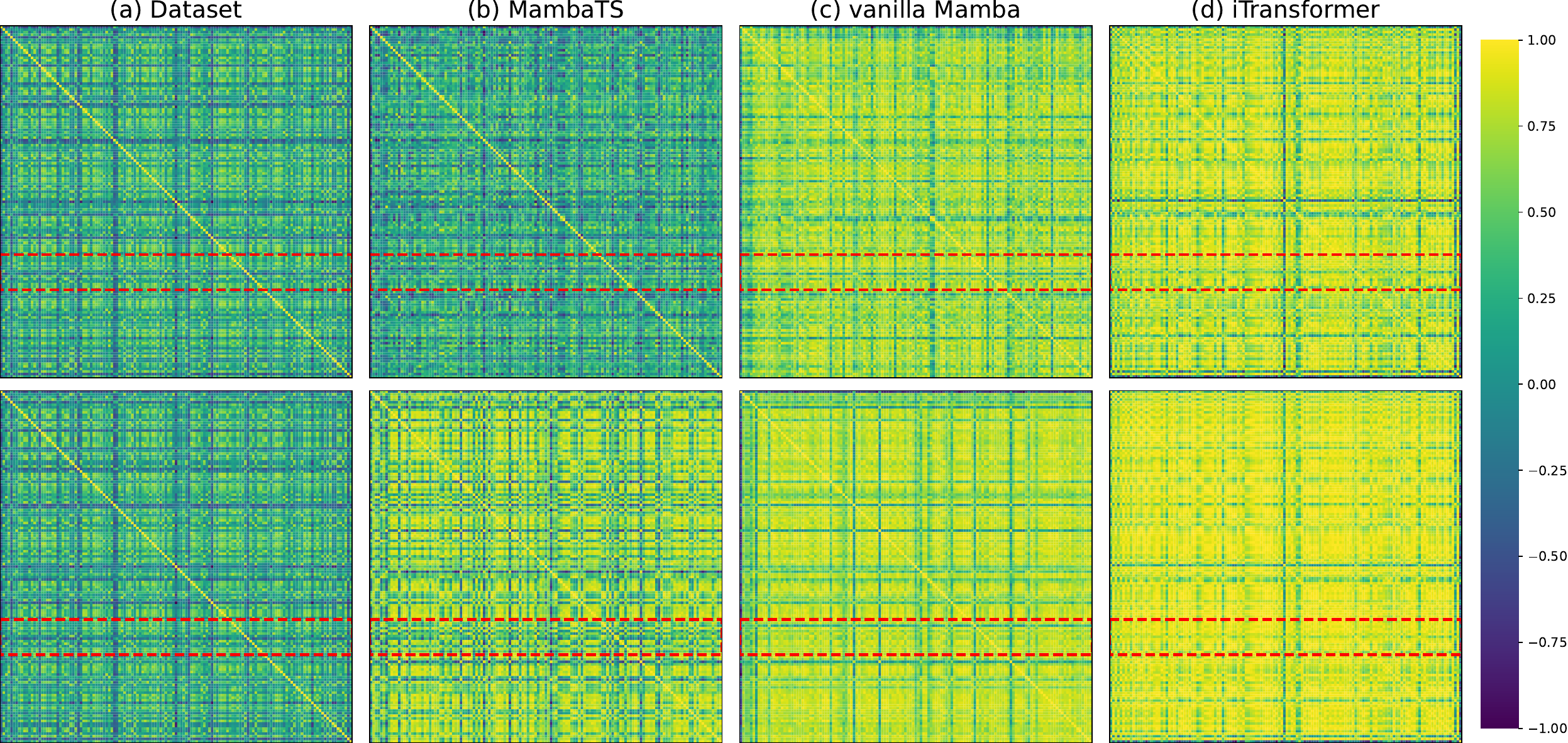}
	\caption{Visualization of variables correlation at different layers, including the first (top row) and final layer (bottom row).}
	\label{fig:heatmap}
\end{figure*}

\paragraph{Dropout Ablation} We further analyzed the role of the dropout in TMB. Figure~\ref{fig:dropout} (left) shows the results of MambaTS with different dropout rates (0.1-0.5) on the Weather dataset. Compared to no dropout, the reduction in MSE of MambaTS increased with the dropout rate, achieving the best performance at 0.2 and 0.3, and performance degradation beyond 0.4. Figure~\ref{fig:dropout} (right) illustrates corresponding loss curves during training, indicating that the dropout in TMB helps prevent premature convergence and overfitting. Additionally, we observed that dropout facilitated lower validation loss.

\paragraph{VAST Ablation} In Table~\ref{tab:vast}, we conducted extensive ablation studies on the VAST strategy, focusing on the design and selection of path decoding strategies. "W/o VPT" in Table~\ref{tab:vast} indicates that MambaTS was trained without VPT as the baseline. "Random (100x)" represents sampling 100 test runs after training with VPT and averaging the results. It can be observed that "Random (100x)" significantly outperformed "W/o VPT", which underscores the effectiveness of VPT. Further visual comparisons in Figure~\ref{fig:vast} show that even random variable scanning outperforms "W/o VPT" in most cases. We then explored different heuristic decoding strategies, including Greedy Strategy (GD), Local Search (LS), Lin and Lernighan (LK), and Simulated Annealing (SA). We defaulted to adopting SA as our solver. As a trade-off between efficiency and performance, we did not employ an exact ATSP solver due to its exponential complexity. As shown in Table~\ref{tab:vast} and Figure~\ref{fig:vast}, SA consistently outperformed other solvers in terms of relative performance consistency. 

Further convergence analysis of the VAST training dynamics is provided in Section~\ref{sec:convergence}

\subsection{Model Analysis}
\label{sec:efficiency}

\begin{table}[!t]
	\centering
	\caption{Computational complexity analysis. SA: Self-attention. Conv: Convolution.}
	\label{tab:analysis_complexity}
	\begin{adjustbox}{max width=\linewidth}
		\begin{tabular}{@{}lccc@{}}
			\toprule
			Method        & \makecell{Temporal\\mixing} & \makecell{Variable\\mixing} & \makecell{Computational\\complexity} \\ 
			\midrule
			Autoformer & SA & MLP  & $\mathcal{O}(L\log L)$  \\ 
			FEDformer  & SA & MLP  & $\mathcal{O}(L)$  \\ 
			MICN  & Conv & Conv & $\mathcal{O}(K^2L)$  \\
			FourierGNN  & GNN & GNN & $\mathcal{O}(KL)$  \\ 
			DLinear  & MLP & -- & $\mathcal{O}(L)$  \\ 
			PatchTST & SA & -- & $\mathcal{O}\left(\left(\frac{L}{P}\right)^2\right)$  \\
			iTransformer & MLP & SA & $\mathcal{O}(K^2)$  \\ 
			MambaTS  & TMB & TMB & $\mathcal{O}\left(\frac{KL}{P}\right)$ \\ 
			\bottomrule
		\end{tabular}
	\end{adjustbox}
\end{table}

%
%

\paragraph{Efficiency Analysis}
MambaTS integrates historical information from all variables using VST and performs global dependency modeling through TMB. The computational complexity of MambaTS is $\mathcal{O}(\frac{KL}{P})$, where $K$ denotes the number of variables, $L$ is the lookback window length, and $P$ is the patch stride.

Compared with leading methods, PatchTST and iTransformer have complexities of $\mathcal{O}\left(\left(\frac{L}{P}\right)^2\right)$ and $\mathcal{O}(K^2)$, respectively. MambaTS achieves a favorable balance between temporal modeling and cross-variable dependency modeling. In our experiments, we set $L=720$ and $P=48$, yielding $M=L/P=15$. On high-dimensional datasets such as Traffic ($K=862$), MambaTS achieves substantially lower complexity than iTransformer, i.e., $\mathcal{O}(KM) \ll \mathcal{O}(K^2)$. While variable-independence-based methods such as DLinear and PatchTST are more computationally efficient, they generally sacrifice inter-variable interactions.

Table~\ref{tab:analysis_complexity} summarizes the complexity of representative baselines. Further runtime results are provided in Section~\ref{sec:runtime}.

\begin{figure*}[!t]
	\centering
	\includegraphics[width=0.90\linewidth]{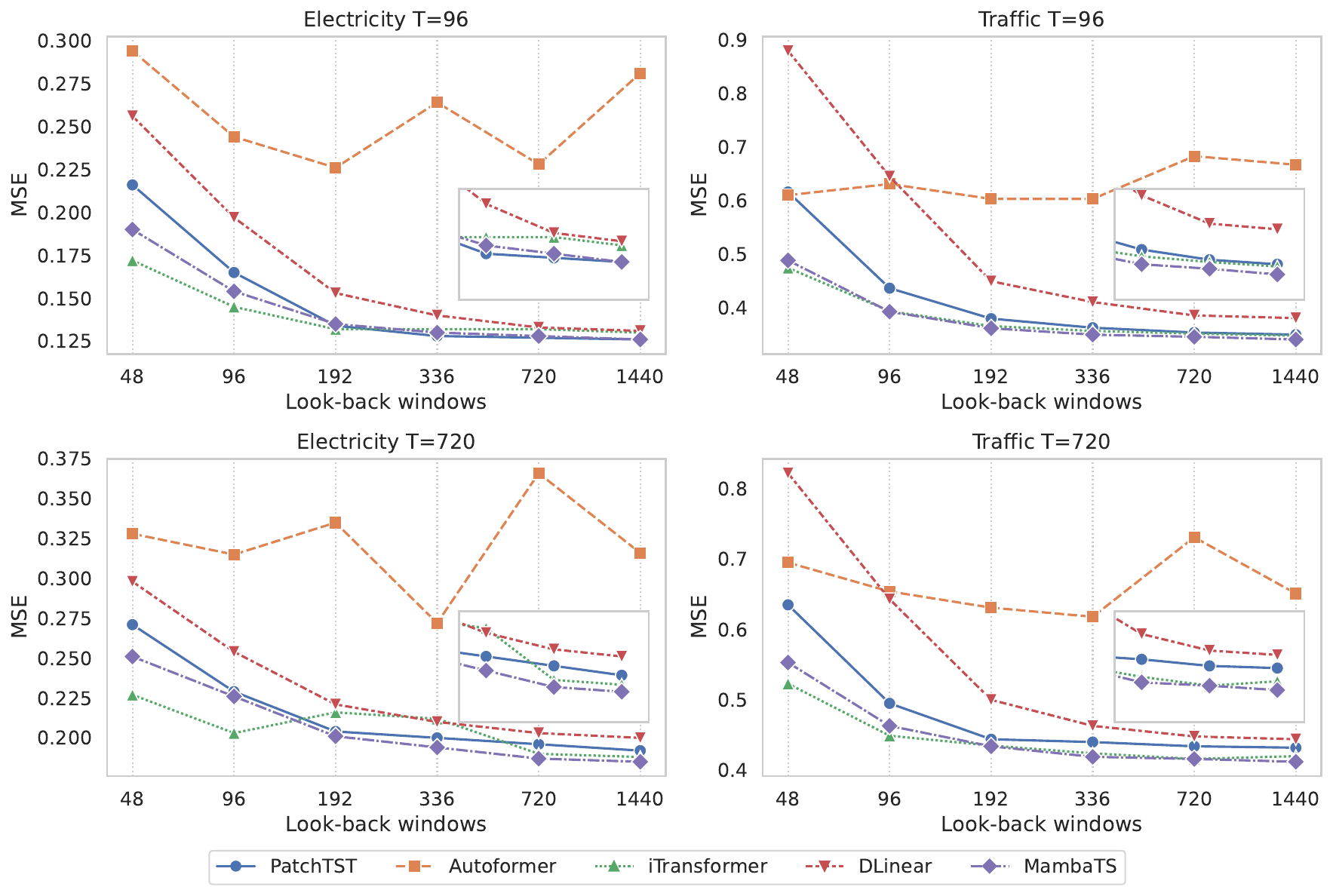}
	\caption{Performance of MambaTS in different datasets with varying length look-back windows.}
	\label{fig:lookback}
\end{figure*}

\paragraph{Variables Dependency Visualization} Compared to the traditional pairwise quadratic complexity of Transformer-based methods, MambaTS estimates the variable dependency graph during training and utilizes this graph for linear scanning during testing, effectively modeling the global dependencies between variables. To further support this claim, we have visualized the dependency graphs learned by MambaTS and iTransformer in Figure~\ref{fig:heatmap}, alongside correlation coefficients between variables to quantify their interdependencies within the dataset. Notably, MambaTS benefits from variable-aware scan along time (VAST), learning richer global dependencies compared to vanilla Mamba (as shown in the red-dashed regions of Figure~\ref{fig:heatmap}). Additionally, while both MambaTS and iTransformer capture similar variable dependencies, MambaTS learns a more intricate dependency graph, whereas iTransformer's graph is smoother, particularly in the final layer, which aligns with the over-smoothing issue often observed in Transformer-based networks~\cite{wu2024demystifying}.

\paragraph{Increasing Lookback Window} Previous studies have shown that Transformer-based methods may not necessarily benefit from a growing lookback window~\cite{Zeng2023Dlinear, nie2023a}, possibly due to distracted attention over the long input. In Figure~\ref{fig:lookback}, we assess MambaTS's performance in this context and compare it with several baselines. It can be observed that MambaTS consistently demonstrates the ability to benefit from the growing input sequence. iTransformer, PatchTST, and DLinear also show this benefit, but MambaTS's overall curve is lower than PatchTST and DLinear. Compared to iTransformer, MambaTS benefits more from a longer lookback window. Additionally, we notice that iTransformer seems to exhibit discontinuous gains on individual dataset tasks.

\section{Conclusion}
\label{conclusion}

We presented MambaTS, a novel framework for long-term multivariate time series forecasting built on improved selective state space models. By integrating structured dependency modeling with a linear scan, MambaTS effectively models global temporal and variable dependencies. The proposed VAST module learns inter-variable relationships during training and determines an optimal scan order at inference. We further introduce the Temporal Mamba Block (TMB) by removing unnecessary causal convolutions from Mamba and incorporate selective dropout to improve generalization. MambaTS demonstrates strong empirical results across eight benchmark datasets, achieving competitive or state-of-the-art performance while maintaining linear complexity.

Despite these merits, several limitations remain. In particular, the scan order in MambaTS is learned offline and remains fixed during inference, which may limit its flexibility in capturing time-varying or non-stationary inter-variable dependencies. In addition, the ATSP-based optimization may introduce scalability concerns in extremely high-dimensional settings, and the learned dependency structure remains implicit rather than directly interpretable. Future work will focus on developing more efficient and potentially adaptive scan-order learning mechanisms, incorporating prior knowledge to enhance interpretability, and extending the framework to more structured and evolving real-world scenarios such as healthcare and spatiotemporal modeling.

We believe that integrating variable relationship discovery with state space modeling provides a promising direction for efficient time series forecasting, with potential impact on both LTSF and the broader SSM community.

\appendix

\section{Additional Comparisons}
\label{sec:additional_exp}
To further evaluate the effectiveness of MambaTS, we additionally compare our method with two recent Mamba-based forecasting models, namely TimeMachine~\cite{ahamed2024timemachine} and Bi-Mamba4TS~\cite{liang2024bi}. TimeMachine adopts a multi-branch Mamba architecture to capture temporal dependencies under different channel interaction patterns and temporal resolutions, while Bi-Mamba4TS combines bidirectional state-space modeling with an adaptive mechanism that dynamically selects between channel-independent and channel-mixing representations. Both methods represent strong recent advances in Mamba-based long-term time series forecasting. Table~\ref{tab:main_sup} reports the complete comparison results.

From the results, MambaTS consistently achieves competitive or superior performance across most datasets and forecasting horizons. In particular, the proposed method demonstrates stronger robustness under long-term forecasting settings, indicating the effectiveness of the designed variable-aware sequential modeling strategy.

While existing Mamba-based forecasting models mainly focus on improving temporal representation learning, MambaTS explicitly investigates the role of variable scan order in multivariate state-space modeling. Our empirical analysis shows that the variable ordering strategy has a substantial impact on forecasting performance, especially for high-dimensional multivariate time series. This observation further validates the motivation behind the proposed architecture design.

Overall, these additional comparisons provide further evidence that MambaTS is a strong and competitive Mamba-based forecasting framework, achieving favorable performance across diverse datasets and prediction horizons.

\section{Runtime Breakdown Analysis}
\label{sec:runtime}

To complement the theoretical complexity analysis in Section~\ref{sec:efficiency}, we further report the practical runtime breakdown of MambaTS, including training, variable-order decoding, and test-time inference. After training, MambaTS determines a variable scanning order by solving an Asymmetric Traveling Salesman Problem (ATSP). Since the obtained scan order is fixed and reused throughout inference, the decoding procedure is executed only once and does not scale with the number of test samples.


Table~\ref{tab:runtime_breakdown} presents the runtime statistics on representative datasets. The results show that the variable-order decoding stage contributes only a small fraction of the total runtime, accounting for 0.01\%, 0.46\%, and 3.12\% of the overall cost on Weather, Solar, and Electricity, respectively. Although the decoding time increases with the number of variables, it remains modest compared with the training cost and is incurred only once after optimization. As a result, the additional overhead introduced by variable-order optimization has a negligible impact on practical deployment.

\begin{figure}[!t]
	\centering
	\includegraphics[width=\linewidth]{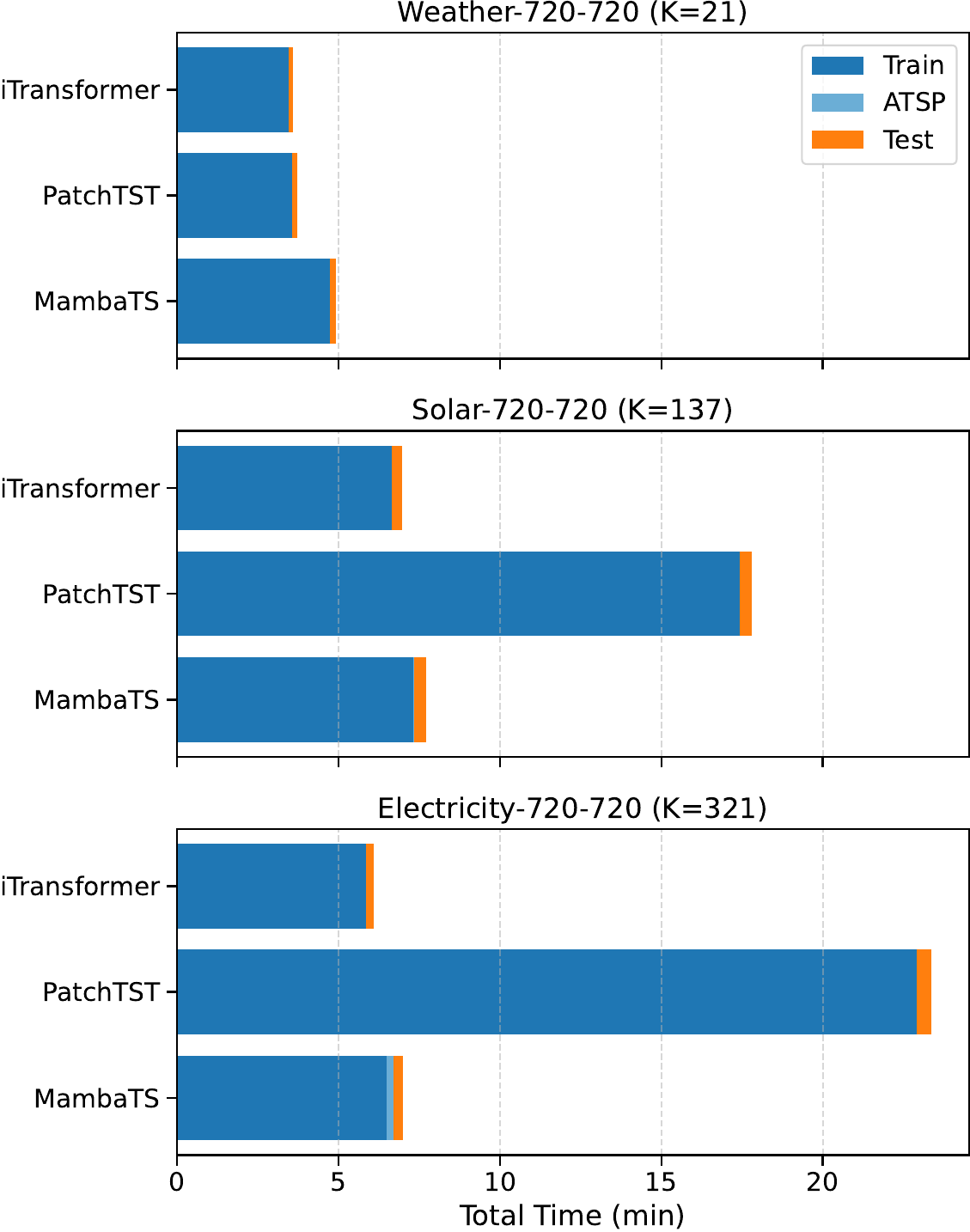}
	\caption{End-to-end runtime comparison of MambaTS, PatchTST, and iTransformer on representative datasets. MambaTS achieves competitive practical efficiency while simultaneously modeling temporal dynamics and cross-variable dependencies.}
	\label{fig:time_breakdown}
\end{figure}

Figure~\ref{fig:time_breakdown} further compares the end-to-end runtime of different forecasting models. Consistent with the theoretical complexity analysis, MambaTS achieves a favorable balance between computational efficiency and modeling capability. While effectively capturing both temporal dynamics and cross-variable dependencies, it maintains competitive runtime performance across datasets with varying numbers of variables. These results demonstrate that the proposed variable-aware design improves forecasting accuracy without introducing substantial computational overhead.

\begin{table}[!t]
	\centering
	\caption{Wall-clock runtime breakdown of MambaTS. The reported runtime includes model training, variable-order decoding via ATSP, and test-time inference. Values in parentheses denote the percentage of total runtime.}
	\label{tab:runtime_breakdown}
	\begin{adjustbox}{max width=\linewidth}
		\begin{tabular}{@{}lccc@{}}
			\toprule
			Dataset & Train & ATSP & Test \\
			\midrule
			Weather ($K=21$) & 283.64 s (96.30\%) & 0.03 s (0.01\%) & 10.88 s (3.69\%) \\
			Solar ($K=137$) & 438.25 s (94.64\%) & 2.14 s (0.46\%) & 22.67 s (4.90\%) \\
			Electricity ($K=321$) & 389.02 s (92.81\%) & 13.06 s (3.12\%) & 17.08 s (4.08\%) \\
			\bottomrule
		\end{tabular}
	\end{adjustbox}
\end{table}

\begin{table*}[!t]
	\caption{Multivariate long-term series forecasting results. All models employ a look-back window length of \(L=96\) for the Covid-19 dataset and \(L=720\) for the remaining datasets. Red bold and blue underlined values denote the best and second-best results, respectively. 
	}
	\label{tab:main_sup}
	\begin{adjustbox}{max width=\linewidth}
		\begin{tabular}{@{}cccccccccccccccccc@{}}
			\toprule
			\multicolumn{2}{c}{Models}                                                   & \multicolumn{2}{c}{ETTh2} & \multicolumn{2}{c}{ETTm2} & \multicolumn{2}{c}{Weather} & \multicolumn{2}{c}{Electricity} & \multicolumn{2}{c}{Traffic} & \multicolumn{2}{c}{Solar} & \multicolumn{2}{c}{Covid-19} & \multicolumn{2}{c}{PEMS} \\ \midrule
			\multicolumn{2}{c}{Metric}                                                   & MSE                                    & MAE                                    & MSE                                   & MAE                                   & MSE                                   & MAE                                  & MSE                                    & MAE                              & MSE                                         & MAE                                & MSE                                       & MAE                                & MSE                                    & MAE                                    & MSE                                   & MAE                                   \\ 
			
			\midrule
			
			\multicolumn{1}{c|}{\multirow{4}{*}{TimeMachine}} & \multicolumn{1}{c|}{96} & \textcolor{red}{\textbf{0.276}} &      \textcolor{red}{\textbf{0.346}} &  \textcolor{blue}{\underline{0.170}} &                                0.266 &  \textcolor{blue}{\underline{0.157}} &                                0.212 &                                0.145 &                                0.240 &                                0.384 &                                0.287 &  \textcolor{blue}{\underline{0.176}} &  \textcolor{blue}{\underline{0.234}} &                                0.935 &                                0.048 &                                0.165 &                                         0.271 \\ 
			\multicolumn{1}{c|}{} & \multicolumn{1}{c|}{192} & \textcolor{blue}{\underline{0.356}} &      \textcolor{red}{\textbf{0.393}} &  \textcolor{blue}{\underline{0.235}} &  \textcolor{blue}{\underline{0.308}} &                                0.205 &                                0.253 &                                0.162 &                                0.254 &                                0.388 &                                0.279 &  \textcolor{blue}{\underline{0.202}} &  \textcolor{blue}{\underline{0.257}} &                                1.339 &                                0.054 &                                0.169 &                                         0.261 \\ 
			\multicolumn{1}{c|}{} & \multicolumn{1}{c|}{336} & \textcolor{blue}{\underline{0.377}} &  \textcolor{blue}{\underline{0.419}} &  \textcolor{blue}{\underline{0.289}} &  \textcolor{blue}{\underline{0.336}} &                                0.260 &                                0.297 &                                0.175 &  \textcolor{blue}{\underline{0.268}} &                                0.390 &                                0.287 &                                0.213 &  \textcolor{blue}{\underline{0.271}} &                                2.666 &                                0.089 &                                0.187 &                                         0.274 \\ 
			\multicolumn{1}{c|}{} & \multicolumn{1}{c|}{720} & \textcolor{red}{\textbf{0.403}} &      \textcolor{red}{\textbf{0.441}} &  \textcolor{blue}{\underline{0.358}} &  \textcolor{blue}{\underline{0.394}} &                                0.318 &                                0.337 &                                0.208 &  \textcolor{blue}{\underline{0.298}} &                                0.435 &                                0.305 &  \textcolor{blue}{\underline{0.218}} &  \textcolor{blue}{\underline{0.269}} &                                7.188 &                                0.152 &                                0.265 &                                0.338 \\ \midrule 
			
			\multicolumn{1}{c|}{\multirow{4}{*}{Bi-Mamba4TS}} & \multicolumn{1}{c|}{96} & 0.311 &                                0.361 &                                0.173 &  \textcolor{blue}{\underline{0.261}} &                                0.159 &  \textcolor{blue}{\underline{0.211}} &  \textcolor{blue}{\underline{0.131}} &  \textcolor{blue}{\underline{0.232}} &  \textcolor{blue}{\underline{0.360}} &  \textcolor{blue}{\underline{0.267}} &                                0.215 &                                0.264 &  \textcolor{blue}{\underline{0.898}} &  \textcolor{blue}{\underline{0.042}} &  \textcolor{blue}{\underline{0.134}} &           \textcolor{blue}{\underline{0.239}} \\ 
			\multicolumn{1}{c|}{} & \multicolumn{1}{c|}{192} & 0.383 &                                0.410 &                                0.237 &      \textcolor{red}{\textbf{0.306}} &  \textcolor{blue}{\underline{0.203}} &  \textcolor{blue}{\underline{0.250}} &  \textcolor{blue}{\underline{0.152}} &  \textcolor{blue}{\underline{0.251}} &  \textcolor{blue}{\underline{0.364}} &  \textcolor{blue}{\underline{0.267}} &                                0.216 &                                0.265 &  \textcolor{blue}{\underline{1.181}} &  \textcolor{blue}{\underline{0.053}} &  \textcolor{blue}{\underline{0.161}} &           \textcolor{blue}{\underline{0.257}} \\ 
			\multicolumn{1}{c|}{} & \multicolumn{1}{c|}{336} & 0.392 &                                0.423 &                                0.326 &                                0.360 &  \textcolor{blue}{\underline{0.252}} &  \textcolor{blue}{\underline{0.287}} &  \textcolor{blue}{\underline{0.170}} &                                0.271 &  \textcolor{blue}{\underline{0.387}} &  \textcolor{blue}{\underline{0.279}} &  \textcolor{blue}{\underline{0.207}} &                                0.280 &  \textcolor{blue}{\underline{2.282}} &  \textcolor{blue}{\underline{0.078}} &  \textcolor{blue}{\underline{0.166}} &           \textcolor{blue}{\underline{0.262}} \\ 
			\multicolumn{1}{c|}{} & \multicolumn{1}{c|}{720} & 0.439 &                                0.464 &                                0.375 &                                0.399 &  \textcolor{blue}{\underline{0.316}} &  \textcolor{blue}{\underline{0.333}} &  \textcolor{blue}{\underline{0.204}} &                                0.300 &  \textcolor{blue}{\underline{0.421}} &  \textcolor{blue}{\underline{0.296}} &                                0.256 &                                0.306 &  \textcolor{blue}{\underline{4.556}} &      \textcolor{red}{\textbf{0.107}} &  \textcolor{blue}{\underline{0.213}} &  \textcolor{blue}{\underline{0.292}} \\ \midrule

			\multicolumn{1}{c|}{\multirow{4}{*}{MambaTS}} & \multicolumn{1}{c|}{96} & \textcolor{blue}{\underline{0.281}} &  \textcolor{blue}{\underline{0.347}} &      \textcolor{red}{\textbf{0.166}} &      \textcolor{red}{\textbf{0.260}} &      \textcolor{red}{\textbf{0.145}} &      \textcolor{red}{\textbf{0.195}} &      \textcolor{red}{\textbf{0.128}} &      \textcolor{red}{\textbf{0.223}} &      \textcolor{red}{\textbf{0.347}} &      \textcolor{red}{\textbf{0.248}} &      \textcolor{red}{\textbf{0.165}} &      \textcolor{red}{\textbf{0.231}} &      \textcolor{red}{\textbf{0.784}} &      \textcolor{red}{\textbf{0.037}} &      \textcolor{red}{\textbf{0.059}} &               \textcolor{red}{\textbf{0.161}} \\ 
			\multicolumn{1}{c|}{} & \multicolumn{1}{c|}{192} & \textcolor{red}{\textbf{0.352}} &  \textcolor{blue}{\underline{0.397}} &      \textcolor{red}{\textbf{0.228}} &      \textcolor{red}{\textbf{0.306}} &      \textcolor{red}{\textbf{0.192}} &      \textcolor{red}{\textbf{0.241}} &      \textcolor{red}{\textbf{0.145}} &      \textcolor{red}{\textbf{0.239}} &      \textcolor{red}{\textbf{0.358}} &      \textcolor{red}{\textbf{0.255}} &      \textcolor{red}{\textbf{0.178}} &      \textcolor{red}{\textbf{0.240}} &      \textcolor{red}{\textbf{1.075}} &      \textcolor{red}{\textbf{0.046}} &      \textcolor{red}{\textbf{0.075}} &               \textcolor{red}{\textbf{0.179}} \\ 
			\multicolumn{1}{c|}{} & \multicolumn{1}{c|}{336} & \textcolor{red}{\textbf{0.372}} &      \textcolor{red}{\textbf{0.416}} &      \textcolor{red}{\textbf{0.276}} &      \textcolor{red}{\textbf{0.335}} &      \textcolor{red}{\textbf{0.245}} &      \textcolor{red}{\textbf{0.280}} &      \textcolor{red}{\textbf{0.163}} &      \textcolor{red}{\textbf{0.259}} &      \textcolor{red}{\textbf{0.372}} &      \textcolor{red}{\textbf{0.262}} &      \textcolor{red}{\textbf{0.192}} &      \textcolor{red}{\textbf{0.252}} &      \textcolor{red}{\textbf{1.776}} &      \textcolor{red}{\textbf{0.064}} &      \textcolor{red}{\textbf{0.102}} &               \textcolor{red}{\textbf{0.206}} \\ 
			\multicolumn{1}{c|}{} & \multicolumn{1}{c|}{720} & \textcolor{blue}{\underline{0.404}} &  \textcolor{blue}{\underline{0.444}} &      \textcolor{red}{\textbf{0.355}} &      \textcolor{red}{\textbf{0.391}} &      \textcolor{red}{\textbf{0.313}} &      \textcolor{red}{\textbf{0.329}} &      \textcolor{red}{\textbf{0.192}} &      \textcolor{red}{\textbf{0.286}} &      \textcolor{red}{\textbf{0.416}} &      \textcolor{red}{\textbf{0.284}} &      \textcolor{red}{\textbf{0.199}} &      \textcolor{red}{\textbf{0.258}} &      \textcolor{red}{\textbf{4.155}} &  \textcolor{blue}{\underline{0.112}} &      \textcolor{red}{\textbf{0.134}} &      \textcolor{red}{\textbf{0.230}} \\
			
			\bottomrule
			\\ 
		\end{tabular}
	\end{adjustbox}
\end{table*}

\section{Convergence Analysis of the Scan-Order Matrix $P$}
\label{sec:convergence}
As described in the manuscript, the matrix $P \in \mathbb{R}^{K \times K}$ is estimated by optimizing the training objective to induce an effective variable scan order. Under the standard assumption that the training and validation sets are independently and identically distributed, a decreasing validation loss indicates that the learned scan order encoded in $P$ is approaching a stable and meaningful configuration. Consequently, stabilization of the validation loss provides an indirect but informative signal for the convergence behavior of $P$. 

To provide explicit evidence of the convergence of $P$, we directly quantify its evolution throughout training. Let $P^{(t)}$ denote the matrix obtained at optimization step (or epoch) $t$. We define the change between consecutive updates as
\begin{equation}
	\Delta P^{(t)} = P^{(t)} - P^{(t-1)}.
\end{equation}

We then track the following complementary metrics.

\textbf{Frobenius norm.}
\begin{equation}
	\|\Delta P^{(t)}\|_F = \sqrt{\sum_{i,j} \left(\Delta P^{(t)}_{ij}\right)^2},
\end{equation}
which measures the overall magnitude of the matrix update.

\textbf{Relative Frobenius difference.}
\begin{equation}
	\mathrm{RelDiff}^{(t)} =
	\frac{\|\Delta P^{(t)}\|_F}{\|P^{(t-1)}\|_F + \epsilon},
\end{equation}
where $\epsilon$ is a small constant for numerical stability. This metric captures convergence independent of the absolute scale of $P$.

\textbf{Entropy difference.}
Since $P$ represents a learned vector-valued dependency structure, we further normalize it row-wise to obtain a stochastic interpretation:
\begin{equation}
	\tilde{P}_{i\cdot}^{(t)} =
	\frac{P_{i\cdot}^{(t)}}{\sum_j P_{ij}^{(t)} + \epsilon}.
\end{equation}

Based on this normalized matrix, we compute the row-wise entropy
\begin{equation}
	H^{(t)} =
	\frac{1}{K} \sum_{i=1}^{K}
	\left(
	-\sum_{j=1}^{K} \tilde{P}_{ij}^{(t)} \log \tilde{P}_{ij}^{(t)}
	\right),
\end{equation}
and define the entropy difference
\begin{equation}
	\Delta H^{(t)} = H^{(t)} - H^{(t-1)},
\end{equation}
which captures distributional changes in the induced scan-order structure.

\textbf{Maximum absolute difference.}
\begin{equation}
	\|\Delta P^{(t)}\|_{\infty} =
	\max_{i,j} \left| \Delta P^{(t)}_{ij} \right|,
\end{equation}
which reflects the largest entry-wise change and ensures that no individual dependency undergoes abrupt variation.

%
%

\begin{figure*}[!t]
	\centering
	\includegraphics[width=1\linewidth]{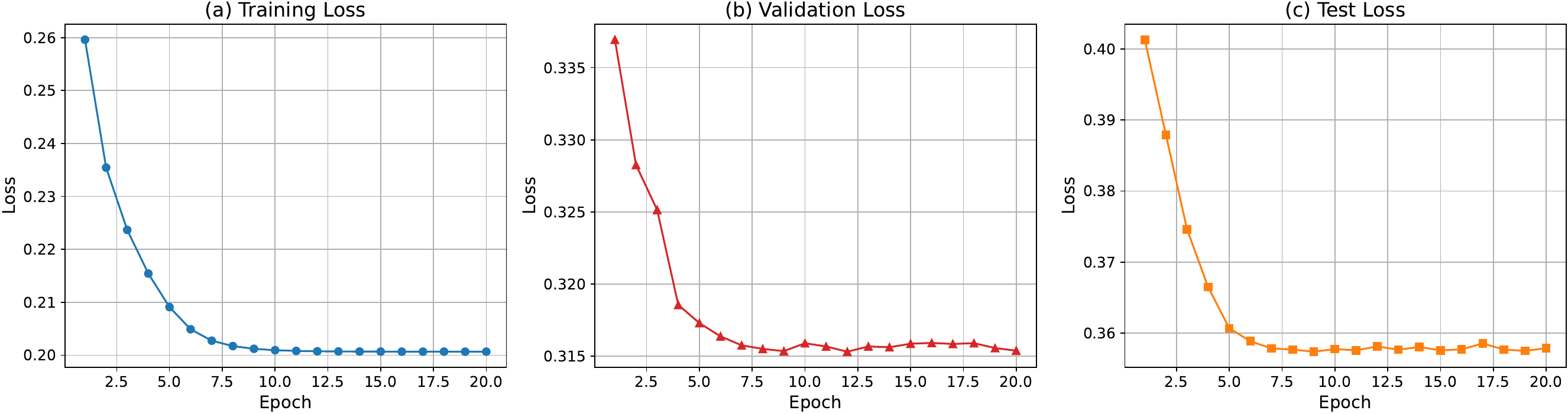}
	\caption{
		Training, validation, and test loss curves during optimization on the Traffic dataset.
	}
	\label{fig:loss_curves}
\end{figure*}

For empirical evaluation, we extend the training duration from the original 10 epochs to 20 epochs to explicitly inspect late-stage structural dynamics. As shown in Fig.~\ref{fig:loss_curves}, the training loss on the Traffic dataset decreases monotonically throughout optimization, while the validation loss reaches its minimum around epoch 9 and subsequently exhibits mild oscillations, suggesting the onset of slight overfitting and a decoupling between training improvement and generalization performance.

\begin{figure*}[!ht]
	\centering
	\includegraphics[width=1\linewidth]{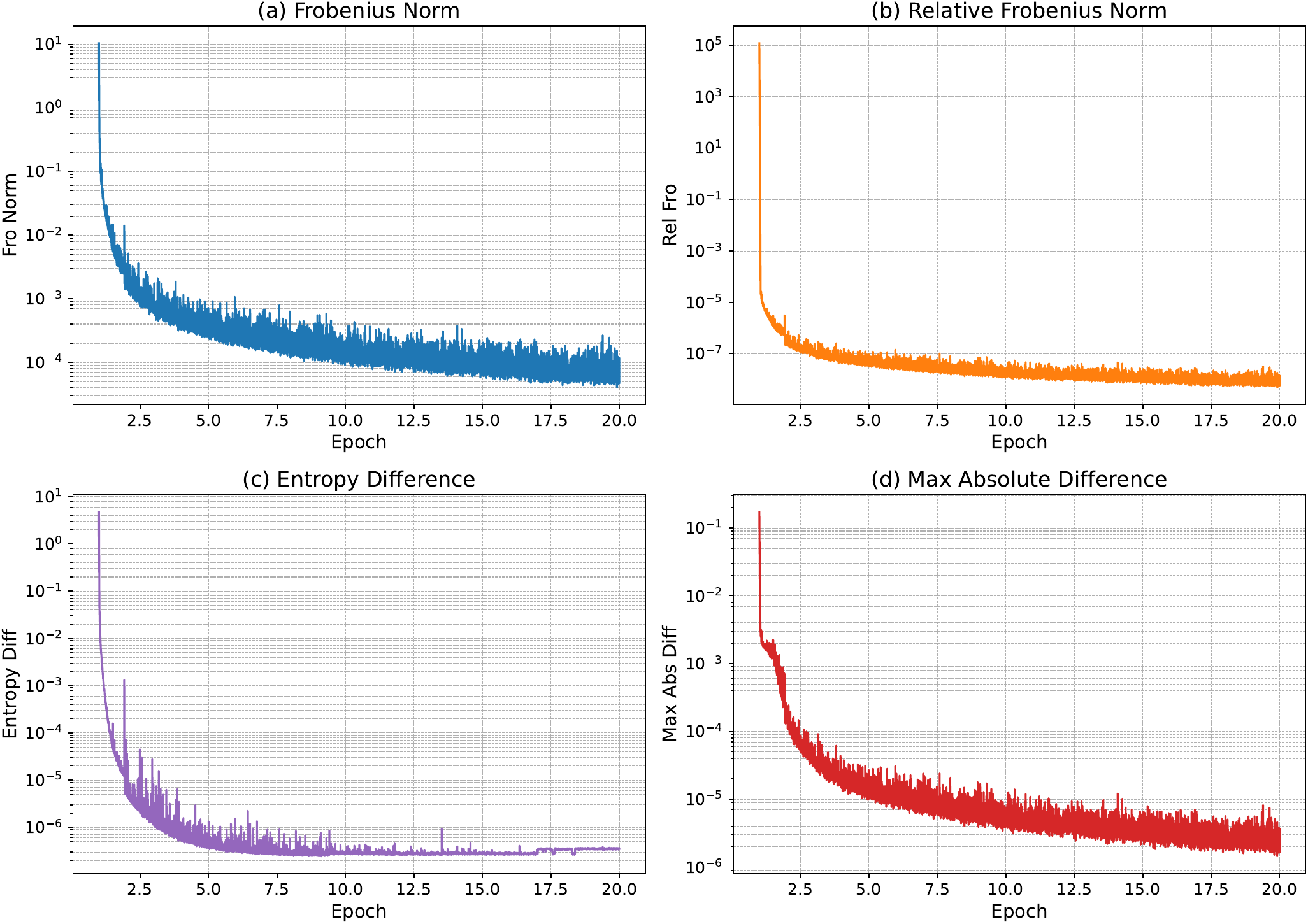}
	\caption{
		Evolution of the scan-order matrix $P$ during training in logarithmic scale. 
		All convergence metrics decrease rapidly in the early stage of optimization and stabilize within a small number of epochs, indicating that the learned scan-order structure converges significantly earlier than training termination.
	}
	\label{fig:p_matrix_log}
\end{figure*}

More importantly, Fig.~\ref{fig:p_matrix_log} reveals a consistent two-stage behavior in the evolution of $P$: a fast adaptation phase followed by a near-stationary regime. All $P$-difference metrics decrease rapidly during early training and stabilize well before optimization terminates. In particular, the relative Frobenius difference falls below $10^{-7}$ around epoch 7.5, and the maximum absolute difference drops below $10^{-5}$ shortly thereafter, indicating that subsequent updates do not significantly alter the learned scan-order structure at either global or entry-wise levels.

Interestingly, the entropy difference reaches its minimum near the same epoch as the validation loss and then exhibits a slight increase. This suggests that late-stage updates tend to affect the sharpness of the row-wise distributions without inducing measurable changes in the overall structural stability of $P$, thereby reflecting a weak sensitivity of entropy-based measures to validation-level fluctuations.

Overall, these results indicate that the matrix $P$ reaches a sufficiently stable configuration within a relatively small number of epochs, even in high-dimensional settings. This provides empirical evidence supporting the robustness and early stabilization of the learned scan-order structure.

\bibliography{main}

\end{document}